\documentclass{article} 
\usepackage{arxiv2027_conference,times}

\usepackage{amsmath,amsfonts,bm}

\def\eqref#1{equation~\ref{#1}}

\def\Algref#1{Algorithm~\ref{#1}}

\def\1{\bm{1}}

\DeclareMathAlphabet{\mathsfit}{\encodingdefault}{\sfdefault}{m}{sl}
\SetMathAlphabet{\mathsfit}{bold}{\encodingdefault}{\sfdefault}{bx}{n}

\usepackage{hyperref}
\usepackage{url}

\usepackage{amsmath}
\usepackage{amssymb}
\usepackage{amsthm}
\usepackage{thmtools}
\usepackage{algorithm}
\usepackage{algorithmicx}
\usepackage[noend]{algpseudocode}

\newtheorem{lemma}{Lemma}
\newtheorem{proposition}{Proposition}

\newtheorem{remark}{Remark}
\newtheorem{theorem}{Theorem}

\usepackage{enumerate}
\usepackage{enumitem}

\usepackage{multirow}
\usepackage{booktabs}

\usepackage[capitalize,noabbrev]{cleveref}

\crefname{ALG@line}{line}{lines}
\Crefname{ALG@line}{Line}{Lines}

\definecolor{cmnistLightBlue}{HTML}{3F68E0}
\definecolor{cmnistBlue}{HTML}{0000FF}
\definecolor{cmnistRed}{HTML}{FD4400}
\definecolor{cmnistBeige}{HTML}{FDDCAC}
\definecolor{shapesBlue}{HTML}{0067FD}
\definecolor{shapesLightGreen}{HTML}{00FC71}
\definecolor{shapesRed}{HTML}{FD0000}
\definecolor{shapesPink}{HTML}{FE009F}
\definecolor{shapesDarkBlue}{HTML}{3E00FE}

\usepackage[dvipsnames]{xcolor}

\usepackage{siunitx}

\usepackage[textsize=tiny]{todonotes}
\usepackage{comment}

\usepackage{bm}
\usepackage{stmaryrd}

\usepackage{environ}

\newif\ifmoveprooftoend
\newcommand\showDeferredProofs{}

\makeatletter
\NewEnviron{movableProof}[2][Proof]{%
        \ifmoveprooftoend
                \edef\next{\noexpand\g@addto@macro\noexpand\showDeferredProofs{%
                        \noexpand\cref{#2}\begin{proof}[#1]\unexpanded\expandafter{\BODY}\noexpand\end{proof}}}
                \next
        \else
                \begin{proof}[#1]\BODY\end{proof}
        \fi}{}
\makeatother
\moveprooftoendtrue

\usepackage[most]{tcolorbox}
\tcbset{
  mytheobox/.style={
    enhanced,
    colback=gray!10,      
    colframe=gray!40,     
    boxrule=0.4pt,
    arc=2pt,
    left=6pt,right=6pt,top=2pt,bottom=2pt,
    before skip=8pt,
    after skip=8pt,
  }
}

\usepackage{empheq}

\usepackage{graphicx}
\usepackage{tikz}
\usetikzlibrary{arrows.meta, positioning, calc, backgrounds, fit}

\usepackage[nolist,nohyperlinks]{acronym}

\usepackage{soul}
\usepackage[normalem]{ulem} 

\definecolor{color1}{HTML}{00F9DE}
\definecolor{color2}{HTML}{C7DFEE}
\definecolor{color3}{HTML}{FADABE}
\definecolor{color4}{HTML}{00008B}
\definecolor{color5}{HTML}{013220}

\newcommand{\method}{\textsc{LoGSI}}

\title{Logical Guidance of Stochastic Interpolants}

\author{%
Gerrit Gerhartz \\
NEC Laboratories Europe\\
ITP, Heidelberg University\\
\And
Francesco Alesiani \\
NEC Laboratories Europe\\
\And
Jonathan Warrell \\
NEC Laboratories America\\
\And
Henrik Christiansen \\
NEC Laboratories Europe\\
\And
Matheus Ferraz \\
NEC Bio B.V. \\
\And
Mathias Niepert \\
NEC Laboratories Europe \\
University of Stuttgart\\
}

\iclrfinalcopy 
\begin{document}

\begin{acronym}
\acro{cvnn}[CVNN]{Complex-Valued Neural Network}
\acro{mlp}[MLP]{Multilayer Perceptron}
\acro{silu}[SiLU]{Sigmoid Linear Unit}
\acro{mae}[MAE]{Mean Absolute Error}
\acro{mse}[MSE]{Mean Square Error}
\acro{ce}[CE]{Cross-Entropy}
\acro{mc}[MC]{Monte Carlo}
\acro{pde}[PDE]{Partial Differential Equation}
\acro{svgd}[SVDG]{Stein Variational Gradient Descent}
\acro{csgd}[CSDG]{Cauchy Schwarz Gradient Descent}
\acro{kl}[KL]{Kullback-Leibler}
\acro{cs}[CS]{Cauchy-Schwarz}
\acro{mcmc}[MCMC]{Markov chain Monte Carlo}
\acro{mc}[MC]{Monte Carlo}
\acro{hmc}[HMC]{Hamiltonian Monte Carlo}
\acro{mh}[MH]{Metropilis-Hasting}
\acro{fm}[FM]{Flow Matching}
\acro{si}[SI]{Stochastic Interpolants}

\end{acronym}

\maketitle

\begin{abstract}
Guidance and composition of generative models have applications in visual design, chemistry, and biology.
Logical guidance composes conditional scores exactly for diffusion models. 
However, \ac{fm} and \ac{si} learn velocity fields.
We show that negation and exclusive disjunction admit exact conditional velocity composition on any interpolant, whereas conjunction, and therefore any formula containing one, is exact on Gaussian paths but in general incurs a drift-interaction error, so composed velocities need not sample the logical target.
\method\ solves this by using score-based re-equilibration as our principled mechanism and falling back to the Gaussian-path velocity rule as an approximate transport predictor.
In non-Gaussian bridge problems, generation alternates transport steps with score re-equilibration steps.
On image transfer (EMNIST to CMNIST, CelebA) and QM9 source-to-target molecular editing with logical attribute constraints, \method\ improves rule conformity at the same network-evaluation budget over composed-velocity baselines, while also improving fidelity on the image tasks.
More broadly, \method\ applies wherever stochastic interpolants need guidance and logical combinations of constraints are natural.
\end{abstract}

\section{Introduction}
Guidance and composition are central to generative AI, enabling controlled generation across domains ranging from visual design to chemistry and biology \citep{saharia2022photorealistic,hoogeboom2022equivariant,Watson2023,skreta:2025,blohmcomposition}. Guidance steers a model toward desired properties, while composition combines multiple conditions or concepts to produce outputs that satisfy more complex requirements. Most existing methods target diffusion models \citep{sohldickstein2015deepunsupervisedlearningusing,ho2020denoising,song2021scorebased,song2020denoising}, where one endpoint is a fixed Gaussian, and the reverse dynamics depend only on the score. A natural extension is to lift logical guidance to stochastic interpolants (SI) \citep{albergo2025stochastic}, which unify diffusion and flow matching \citep{lipmanflow} and are not tied to a Gaussian endpoint. 
Classical guidance \citep{dhariwal2021diffusion,ho2022classifier} for SI has been studied \citep{feng2025guidance,jiang2026guided}, but logical composition of atomic conditions in this regime remains open. We focus on domain transfer \citep{liu20232}: an interpolant transports a source distribution $\rho_0$ to a target $\rho_1$, samples can be initialized from concrete source instances \citep{albergo2024stochasticcouplings}, and guidance enforces a Boolean formula $\varphi$ over target attributes. 
For example, we may edit a source molecule by adding a geometry (e.g.\ rings) and an atom-composition constraint (e.g.\ oxygen).

Throughout, we distinguish between unconditional velocity and score fields (or velocities and scores for short), denoted by $v_t(x)$ and $s_t(x)$, and conditional velocity and score fields, denoted by $v_t(x\mid c)$ and $s_t(x\mid c)$, which steer generation toward samples satisfying a constraint or property $c$.

The core challenge is that SIs learn conditional velocities, but it is not clear a priori how these fields should be combined to enforce a Boolean formula $\varphi$. For conditional scores, 
\citet{alesiani2026logical} (LogDiff) derive exact recursive composition rules. We show that the analogous velocity composition is exact for Gaussian-path bridges, and that outside that regime it fails if the Boolean formula contains conjunctions or conditional-independent disjunctions (CI-disjunctions).

\begin{figure}[t]
    \centering
    \includegraphics[width=1.0\textwidth]{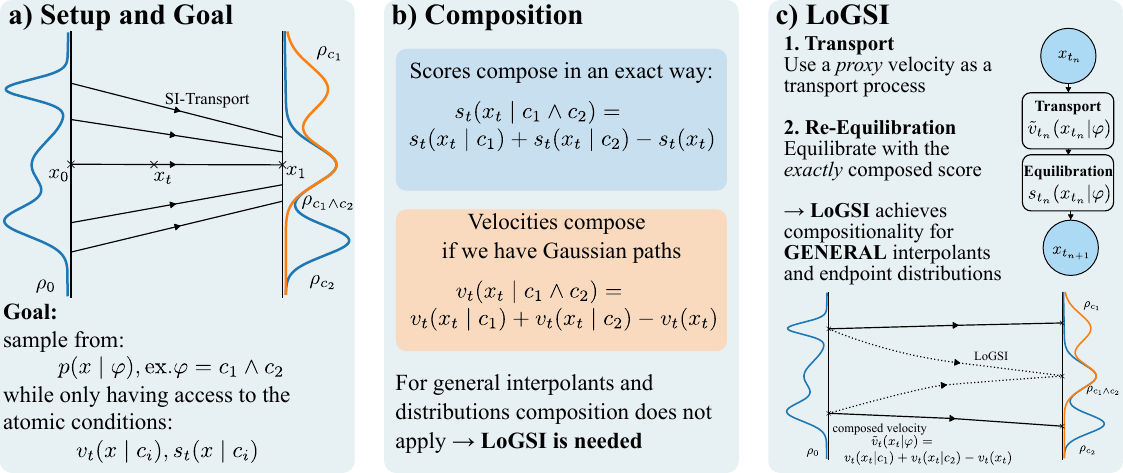}
    \caption{%
        \textbf{Overview of \method.}
        \textbf{(a)}~Logical composition targets samples from $\rho_1$ satisfying $c_1 \land c_2$, transported from $\rho_0$ by a stochastic interpolant, sampling can only use the atomic conditioned scores and velocities.
        \textbf{(b)}~Scores compose exactly for general stochastic interpolants, velocity composition matches the same rule on Gaussian-path / diffusion bridges, except at negation and exclusive-disjunction nodes, which match on any interpolant (\cref{thm:not-orme-exact}).
        \textbf{(c)}~Our method transports samples with a proxy velocity roughly pointing in the right direction, then re-equilibrates with the exactly composed conditional score.
    }
    \label{fig:overview}
\end{figure}

We introduce \method\ to address this asymmetry between velocity and score compositionality (\cref{fig:overview}). On general stochastic interpolants with a non-Gaussian starting distribution, both velocity and score are available at sampling time, but only the score admits exact logical composition for every formula~\citep{alesiani2026logical}. In \method, composed \emph{velocities} are exact on Gaussian-path bridges and can be used as a \emph{proxy} transport field elsewhere. \method\ therefore separates roles: an approximate transport step along the \emph{proxy} transport field $\tilde v_t(\cdot\mid\varphi)$ advances mass along the interpolant, while exact composed conditional scores re-equilibrate samples toward $p_t(\cdot\mid\varphi)$ at each timestep. In the limit of sufficient score-correction steps at fixed $t$, the procedure is consistent with the logical target. Concretely, we realize this split as a transport--re-equilibration sampler: each outer step applies a transport update along $\tilde v_t$, followed by $K$ overdamped Langevin steps with the exactly composed conditional score.

We evaluate our method on image and molecular transfer with non-Gaussian endpoints, testing general Boolean templates (conjunction, disjunction, negation) and randomly generated expression trees. At matched network-evaluation budget, score-guided re-equilibration with a proxy velocity improves rule conformity over composed-velocity integration.  It also yields higher fidelity on the image experiments. 

Our main contributions are as follows:
\begin{itemize}
        \item \textbf{Score--velocity asymmetry for logical SI.} On Gaussian-path flow matching, LogDiff score composition induces an exact velocity composition rule. For general stochastic interpolants with non-Gaussian endpoints, score composition carries over unchanged, velocity rules fail for conjunction and CI-disjunction, and therefore for general logical statements $\varphi$.
        \item \textbf{Logical guidance by score correction.} We sample under Boolean formulas by re-equilibrating with exactly composed conditional scores at each timestep, avoiding any closed-form composed velocity field.
        \item \textbf{Proxy velocity transport.} We reuse the Gaussian-path velocity rule as a transport proxy, cutting the number of correction steps needed and improving conformity over the composed-velocity baseline.
\end{itemize}

\section{Related Work}

\textbf{Generative methods and bridges.}
Diffusion models generate data by reversing a fixed noising process \citep{sohldickstein2015deepunsupervisedlearningusing, ho2020denoising, song2020generativemodelingestimatinggradients}, unified by \citet{song2021scorebased} in the notion of stochastic processes. \citet{song2021scorebased} also propose to sample with a predictor-corrector scheme, where a Langevin corrector uses the score to remove the predictor's discretization error while both target the same marginal. Our sampler is similar but structurally different: because only the score composes exactly for formulas containing conjunctions or general disjunctions (\cref{rem:score-exact}), the score--velocity asymmetry (\cref{sec:vel-obstruction}) forces the transport step to be a deliberate proxy $\tilde v_t\neq v_t(\cdot\mid\varphi)$, while the corrector is the exactness mechanism enforcing the logical constraint, needed beyond discretization error even with infinitesimal transport steps. Flow matching \citep{lipmanflow} learns a velocity field between two distributions, and stochastic interpolants \citep{albergo2025stochastic} generalize both diffusion and flow matching while dropping the Gaussian-endpoint assumption, which is why we adopt them for our method. Finally, Schrödinger bridges \citep{de2021diffusion, shi2023diffusion} transport between arbitrary distributions under an entropy-optimal-transport objective. Since our logical composition is orthogonal to how the bridge is trained, extending it to such samplers is an appealing direction for future work.

\textbf{Guidance and compositional conditioning.}
Guidance of generative models was first achieved by training a separate classifier to steer diffusion sampling \citep{dhariwal2021diffusion}. The noise-aware classifiers we use for the posterior weights in our logical composition play an analogous role. Classifier-free guidance \citep{ho2022classifier} shortly after removed the need for a separately trained classifier by jointly modeling conditional and unconditional scores. Both approaches condition on a single condition and do not address how multiple conditions should be combined. Compositional generation, started in products of experts \citep{hinton:2002} and revived more recently \citep{du:2023, gaudi:2025}, targets exactly this problem, and subsequent work developed compositional and logical conditioning for diffusion models \citep{liu2022compositional}. Closely related, \citet{skreta:2025} introduce SuperDiff, which composes pretrained diffusion models at inference time through logical \texttt{AND} and \texttt{OR} operators. We discuss the connection to our bridge setting in \cref{app:skreta}. Most relevant to us, \citet{alesiani2026logical} derive exact recursive rules for composing conditional scores under Boolean formulas. Guidance has also been extended to flow matching \citep{feng2025guidance} and stochastic interpolants \citep{jiang2026guided}, but to our knowledge logical composition in these settings remains unexplored.

\section{Preliminaries}
To build our method we rely on the framework of stochastic interpolants by \cite{albergo2025stochastic} and the logical guidance work of \cite{alesiani2026logical}.

\subsection{Stochastic Interpolants}
\label{sec:prelim-si}
In this work we use the framework of stochastic interpolants by \cite{albergo2025stochastic}, which unifies diffusion and flow matching models. The idea of this framework is to replace the diffusion or linear interpolation of diffusion and flow matching models between the two distributions $\rho_0$ and $\rho_1$ with an interpolant of the form
\begin{align}
        x_t = I(t, x_0, x_1)+\gamma(t)z,\quad t\in[0, 1],\,z\sim \mathcal N (0, I),
\end{align}
where $(x_0,x_1)\sim \pi$ is a coupling with marginals $x_0\sim \rho_0$ and $x_1\sim \rho_1$, the interpolation $I$ satisfies the boundary conditions $I(0, x_0, x_1)=x_0$ and $I(1, x_0, x_1)=x_1$, and the noise scheduler $\gamma$ satisfies $\gamma(0)=\gamma(1)=0$ and $\gamma(t)>0$ for all $t\in(0,1)$. One can think of this as an interpolation between the distributions $\rho_0$ and $\rho_1$, bridged by intermittently adding Gaussian noise to the process.
The cases $\gamma\equiv 0$ (flow matching) and Gaussian-path diffusion are special cases~\citep{albergo2025stochastic}. We work with general endpoints $\rho_0,\rho_1$ and refer to \cref{sec:gaussian-path} for the Gaussian-path theory used in our velocity composition result.
In the bridge problem we regress two targets between the given target distributions: the velocity generated by the interpolation $I$ and the score generated by the noise scheduler $\gamma$. The training objectives are
\begin{align}
        \mathcal L_{v}=\mathbb E\bigl\|v_{t}(x_t)-(\partial_t I(t, x_0, x_1)+\dot{\gamma}(t)z)\bigr\|^2,\quad \mathcal L_{s}=\mathbb E\bigl\|s_{t}(x_t)+\tfrac{z}{\gamma(t)}\bigr\|^2,
\end{align}
where the expectations are over $t$, $(x_0,x_1)\sim\pi$, and $z\sim\mathcal N(0,I)$. The minimisers define the marginal velocity and score fields
\begin{align}
        v_t(x) &= \mathbb E\bigl[\partial_t I(t, x_0, x_1)+\dot{\gamma}(t)z\mid x_t=x\bigr],\quad s_t(x) = \nabla_x \log p_t(x),
\end{align}
where $p_t$ is the probability density of $x_t$. For a later derivation we will also need that, by Tweedie's identity, the Gaussian noise component implies that $\mathbb E[z\mid x_t=x] = -\gamma(t)\,s_t(x).$
With these two fields trained, we can sample the stochastic interpolant deterministically (ODE) or stochastically (SDE).
For ODE-based sampling only the velocity is needed:
\begin{align}
        \mathrm dx_t=v_{t}(x_t)\mathrm d t, \quad x_0\sim\rho_0.
\end{align}
For SDE-based sampling the score is also needed explicitly to control the amount of injected noise:
\begin{align}
        \mathrm d x_t = \Bigl(v_{t}(x_t)+\tfrac{1}{2}\bar{\gamma}(t)s_{t}(x_t)\Bigr)\mathrm d t+\sqrt{\bar{\gamma}(t)}\,\mathrm d W_t, \quad x_0\sim\rho_0,
\end{align}
where $\bar{\gamma}(t)\ge 0$ is a sampling noise scheduler, which is distinct from the training scheduler $\gamma$. Empirically choosing them to be equal $\gamma=\bar \gamma$, works well. All SDEs, regardless of the noise scheduler, share the same marginal $p_t$ as the probability-flow ODE.

\subsection{Exact compositions of scores}
\label{sec:prelim-logdiff}
For diffusion models, exact logical composition was developed by \cite{alesiani2026logical}. There only the score enters the reverse-time dynamics, so it is natural to work with posterior logical scores rather than with conditional densities directly.
Let $\mathcal C=\{c_1,\ldots,c_n\}$ be a set of atomic predicates (e.g. atomic conditions), and let $\varphi$ be a Boolean formula over $\mathcal C$ built from conjunction, disjunction, and negation. For $t\in(0,1]$ and $x\in\mathbb R^d$, define the backward truth probability and the logical score
\begin{align}
        p_t(\varphi\mid x) &:= \mathbb P\bigl(x_1\models \varphi \mid x_t=x\bigr),\quad s_t(\varphi\mid x) := \nabla_x \log p_t(\varphi\mid x),
\end{align}
where $x_1\models\varphi$ means that the terminal state satisfies $\varphi$.
By Bayes' rule, the conditional score, used for generation, satisfies
$s_t(x\mid \varphi):=\nabla_x \log p_t(x\mid \varphi)=s_t(x)+s_t(\varphi\mid x)$.
For an atomic predicate $c_i$, the corresponding logical score can be obtained from a conditional and an unconditional score network as $s_t(c_i\mid x)=s_t(x\mid c_i)-s_t(x)$.
Exact composition is possible whenever $\varphi$ admits a probabilistic-circuit representation whose conjunction nodes combine conditionally independent subformulas and whose disjunction nodes combine conditionally independent or mutually exclusive subformulas \citep{alesiani2026logical}. The exact recursive formulas are given in Appendix~\ref{app:recursive_scores}. Under these conditions, the composed conditional score is an affine combination of the atomic conditional scores: letting $k$ denote the number of atoms $c_1,\ldots,c_k$ appearing in $\varphi$, there exist coefficients $\alpha_i^{\varphi}(t,x)$, determined recursively from the posterior estimates $\widehat\pi(\cdot):=p_t(\cdot\mid x)$, such that
\begin{align}\label{eq:logdiff_lin_comb_score_xphi}
        s_t(x\mid \varphi)
        = s_t(x)+\sum_{i=1}^{k}\alpha_i^{\varphi}(t,x)\bigl[s_t(x\mid c_i)-s_t(x)\bigr].
\end{align}
In practice the atomic posteriors $\widehat\pi(c_i)= p_t(c_i\mid x)$ are estimated with a noise- and time-aware classifier, while the score terms come from the trained conditional and unconditional networks.

\section{Method}
\label{sec:method}
We first characterize when logical composition extends from scores to velocities (\cref{sec:gaussian-path}), identify exactly where this extension fails for general stochastic interpolants (\cref{sec:vel-obstruction}), and then describe our sampling method (\cref{sec:pc-scheme,sec:proxy-vel}) which resolves the problems.

\subsection{Gaussian-path logical composition}
\label{sec:gaussian-path}

We generalize LogDiff-style guidance \citep{alesiani2026logical} to Gaussian-path flow matching, where the source distribution $\rho_0$ is Gaussian, and show that velocity composition in the sense of LogDiff is then exact (\cref{prop:vel-compose}). Throughout, we take $x_t=\alpha_t x_1+\beta_t\epsilon,\, \epsilon\sim\mathcal N(0,I)$. On such a path the conditional velocity is an affine function of the conditional score, $v_t(x\mid c)=a_t x+b_t\,s_t(x\mid c)$, with coefficients $(a_t,b_t)$ that depend on $t$ but not on the conditioning. This is true for every terminal condition $c$ determined by $x_1$ whose noising kernel is independent of $c$ given $x_1$. This relation is standard, see e.g.\ \citet{zheng2023guided}. We restate it with the explicit coefficients as \cref{lem:affine-vs} in Appendix~\ref{sec:proof-gp-fm}. Combining it with the compositionality of the scores gives the logical composition rule for velocities. (\cref{eq:logdiff_lin_comb_score_xphi})
\begin{proposition}[Exact logical velocity composition on Gaussian paths]
\label{prop:vel-compose}
Let $\varphi$ be a Boolean formula whose score satisfies the linear combination form of \cref{sec:prelim-logdiff} and $\alpha_i^{\varphi}$ the coefficients of LogDiff. On the Gaussian path above,
\begin{align}
        v_t(x\mid\varphi)=v_t(x)+\sum_{i=1}^{k}\alpha_i^{\varphi}(t,x)\bigl[v_t(x\mid c_i)-v_t(x)\bigr].
        \label{eq:gp-vel-compose}
\end{align}
\end{proposition}

This is a sufficiency result: because $(a_t,b_t)$ are independent of the conditioning, the composed score of \cref{sec:prelim-logdiff} induces the matching composed velocity, whereas necessity of the affine relation is not established. Proofs are in \cref{sec:proof-gp-fm} and the recursive formulas for the $\alpha_i^{\varphi}(t,x)$ in \cref{app:recursive_vel}. 
For general stochastic interpolants this rule becomes invalid in general, while negation and exclusive disjunction remain exact, as we show next.

\subsection{Velocity composition on general stochastic interpolants}
\label{sec:vel-obstruction}

We now drop the Gaussian-path assumption and return to the general stochastic-interpolant framework of \cref{sec:prelim-si}: $(x_0,x_1)\sim\pi$ with $x_0\sim\rho_0$, $x_1\sim\rho_1$ (both possibly non-Gaussian), and
\begin{align}
        x_t=I(t,x_0,x_1)+\gamma(t)z,\qquad z\sim\mathcal N(0,I),
\end{align}
with the same boundary conditions on $I$ and $\gamma$ as in the preliminaries. 
As in \cref{sec:gaussian-path}, we seek a relation between the conditional velocity and score. For a general stochastic interpolant, the marginal velocity admits the decomposition
\begin{align} v_t(x\mid c)=\mu_t(x\mid c)-\dot\gamma(t)\gamma(t)s_t(x\mid c), \qquad \mu_t(x\mid c)=\mathbb E\bigl[\partial_t I(t,x_0,x_1)\mid x_t=x,c\bigr], \label{eq:general-vel-decomp}
\end{align}
which separates a conditioning-dependent drift $\mu_t(x\mid c)$ from a score term whose coefficient depends on $t$ only. This identity is established in the theoretical analysis of \citet{albergo2025stochastic}.
It is then natural to test whether LogDiff composition, exact for scores, remains exact when applied directly to velocities. We define $\check v_t(x\mid\varphi)$ and $ \check \mu_t(x\mid\varphi)$ as
\begin{align}
        \check \phi_t(x\mid\varphi)
        :=\phi_t(x)+\sum_{i=1}^{k}\alpha_i^{\varphi}(t,x)\,\left[\phi_t(x\mid c_i)-\phi_t(x)\right],\quad \phi_t\in\{v_t,\mu_t\}.
        \label{eq:partition-vel}
\end{align}
Since the score term in \cref{eq:general-vel-decomp} composes exactly, i.e.  $\check s_t(x\mid\varphi) = s_t(x\mid\varphi), \forall \varphi$ (under the assumptions given in \cref{sec:prelim-logdiff}), 
the question of when \cref{eq:partition-vel} is exact reduces entirely to the drift: the two fields differ exactly by $\check\mu_t(x\mid\varphi)-\mu_t(x\mid\varphi)$.
By writing $\mathbf 1_\psi:=\mathbf 1\{x_1\models\psi\}$, we define the logical drift moment of a subformula $\psi$ as
\begin{align}
        m^{\psi}_t(x):=\mathbb E\bigl[\partial_t I(t,x_0,x_1)\,\mathbf 1_\psi \bigm| x_t=x\bigr],
        \qquad\text{so that}\qquad
        \mu_t(x\mid\psi)=\frac{m^{\psi}_t(x)}{\widehat\pi(\psi)}.
        \label{eq:drift-moment}
\end{align}
Both $m^{\psi}_t$ and $\widehat\pi(\psi)$ are linear in the indicator $\mathbf 1_\psi$, so every Boolean operation is inherited exactly except conjunction (and therefore also for CI-disjunction, since $c_1\lor c_2=\neg(\neg c_1\land\neg c_2)$), where $m^{\psi\land\chi}_t=\mathbb E[\partial_t I\,\mathbf 1_\psi\mathbf 1_\chi\mid x_t=x]$ weights the drift by a product of indicators that does not reduce to the atomic moments (see \cref{lem:moment-algebra} in the Appendix).
Two consequences follow, one positive (\cref{thm:not-orme-exact}) and one negative (\cref{thm:and-bias}).

\begin{theorem}[Exact velocity composition for negation and mutual exclusive disjunction]
\label{thm:not-orme-exact}
For any interpolant $I$, noise schedule $\gamma$, coupling $\pi$ and endpoints $\rho_0,\rho_1$, the NOT rule (whenever $\widehat\pi(\psi)\in(0,1)$) and the OR-ME rule (whenever $\mathbf 1_\psi\mathbf 1_\chi\equiv0$) of \cref{app:recursive_vel} are exact for velocities. Consequently, if $\varphi$ admits a probabilistic-circuit representation containing no conjunction or CI-disjunction node, then $\check v_t(\cdot\mid\varphi)=v_t(\cdot\mid\varphi)$ on every stochastic interpolant.
\end{theorem}

For conjunction the obstruction can be computed in closed form, and it turns out to be an interaction between the atomic conditions.

\begin{theorem}[Exact obstruction for conjunction]
\label{thm:and-bias}
Let $\varphi=c_1\land c_2$, assume the atoms are conditionally independent given $x_t$ in a neighbourhood of $x$ with $\widehat\pi(c_i)\in(0,1)$, so that LogDiff's AND rule ($\alpha^{\varphi}_1=\alpha^{\varphi}_2=1$) gives the exact composed score and $\widehat\pi(\varphi)=\widehat\pi(c_1)\widehat\pi(c_2)>0$, and let $d_t(a_1,a_2):=\mu_t(x\mid c_1^{a_1}\land c_2^{a_2})$, $a\in\{0,1\}^2$, be the four cell-conditional drifts, where $c^1=c$ and $c^0=\lnot c$. Then, for every interpolant $I$, noise schedule $\gamma$, coupling $\pi$ and endpoints $\rho_0,\rho_1$,
\begin{align}
        \check v_t(x\mid\varphi)-v_t(x\mid\varphi)
        &=-\bigl(1-\widehat\pi(c_1)\bigr)\bigl(1-\widehat\pi(c_2)\bigr)\,\Delta_t(x),
        \label{eq:and-bias-cell}\\
        \Delta_t(x)&:=d_t(1,1)-d_t(1,0)-d_t(0,1)+d_t(0,0).
        \label{eq:delta-def}
\end{align}
The bias thus factorises into a purely probabilistic prefactor and the drift interaction $\Delta_t$. Composition at a conjunction is therefore exact if and only if $\bigl(1-\widehat\pi(c_1)\bigr)\bigl(1-\widehat\pi(c_2)\bigr)\Delta_t(x)=0$, in particular whenever the conditional drift is additive $d_t(a_1,a_2)=\theta_0+\theta_1a_1+\theta_2a_2$ in the truth values of the atoms, $\Delta_t(x)=0$, and asymptotically as $\widehat\pi(c_i)\to1$.

\end{theorem}
The same computation applies at any conjunction node with conditionally independent children, so each such node contributes a residual of the form \cref{eq:delta-def}. Composed velocities are exact only if all of these vanish or cancel, which does not hold in general, as the counterexample in \cref{par:explicit-counterexample} shows. The same applies to CI-disjunction through $\psi \vee \xi = \neg(\neg\psi \wedge \neg\xi)$. Proofs are in Appendix~\ref{sec:proof-vel-obstruction}. \cref{rem:gp-vanishing} shows that Gaussian-path structure forces $\Delta_t\equiv0$. 
Since a logical statement containing conjunctions cannot be used to combine velocities, we need a sampler that still enforces conformity, even when the velocity composition is not exact.

\begin{remark}[Scores still compose exactly]
\label{rem:score-exact}
The obstruction in \cref{thm:and-bias} is specific to general composed velocities. The composed conditional score $s_t(x\mid\varphi)$ from \cref{sec:prelim-logdiff} remains exact on general stochastic interpolants under the same conditional-independence and exclusivity assumptions as in LogDiff.
\end{remark}

\subsection{Guiding general stochastic interpolants}
\label{sec:pc-scheme}

In the general bridge problem, the probability-flow ODE and its equivalent SDE family share the marginals $p_t$, but only the score can be composed exactly for general logical statements $\varphi$ (\cref{rem:score-exact}). We therefore split sampling into two roles: (i) a {\it transport} step advances mass toward $\rho_1$, and (ii) exact composed scores {\it re-equilibrate} each sample to the correct conditional marginal $p_t(\cdot\mid\varphi)$.

\textbf{Transport-re-equilibration.}
On a time grid $0=t_0<\cdots<t_N=1$ with step $\Delta t_n=t_{n+1}-t_n$, each outer iteration first takes a single transport step along $\tilde v_t(\cdot\mid\varphi)$, using either the ODE or SDE transport (Line 9 and 11 of \cref{alg:logsi}, a hyperparameter of our method). It then runs $K\ge 0$ overdamped Langevin correctors at fixed $t=t_{n+1}$ after transport,
\begin{align}
        x_{n+1}^{(k+1)} = x_{n+1}^{(k)}
  + \tfrac{1}{2}\bar\gamma(t_{n+1})\,s_{t_{n+1}}(x_{n+1}^{(k)}\mid\varphi)\,\Delta t_n
  + \sqrt{\bar\gamma(t_{n+1})\,\Delta t_n}\,\eta_n^{(k)},
        \label{eq:corrector}
\end{align}
initialized at the transported state, with $\eta_n^{(k)}\sim\mathcal N(0,I)$ independent, for $k=0,\ldots,K-1$. As in diffusion sampling, correctors are not applied at the endpoints, where the score is undefined.
We have picked $\tfrac{1}{2}\gamma(t_{n+1})\Delta t_n$ as Langevin step size to not introduce a new hyperparameter. In principle one can choose other step sizes, and we discuss this in Appendix~\ref{par:langevin_step_size}.
The conditional score is assembled from the atomic conditional scores 
via \cref{sec:prelim-logdiff}. For an algorithmic depiction, see \Algref{alg:logsi}, \cref{alg:compose-weights} and \cref{alg:compose-score} in the appendix. For an discussion about the consistency of this equilibration see \cref{rem:corrector-invariance}.

\subsection{Proxy velocity for transport}
\label{sec:proxy-vel}

A natural, easy to compute transport step uses the unconditional velocity, $\tilde v_t(x\mid\varphi):=v_t(x)$.
However, this step advances the sample along trajectories that are typically misaligned with $p_t(\cdot\mid\varphi)$, therefore requiring many Langevin equilibration steps to re-equilibrate with the logical marginal at each noise level.
We instead propose to use the Gaussian-path composition rule \cref{eq:partition-vel} as a \emph{proxy} velocity $\tilde v_t(x\mid\varphi)=\check v_t(x\mid\varphi)$,
evaluated with the atomic conditional velocities of the trained stochastic interpolant. Outside the Gaussian-path regime this field is not equal to the true conditional velocity $v_t(x\mid\varphi)$ in general (\cref{thm:and-bias}), but it is the unique velocity that would be correct if the affine relation of \cref{lem:affine-vs} held. Additionally, by \cref{thm:not-orme-exact} it is exact at every negation and exclusive-disjunction node of $\varphi$, which is what makes \cref{eq:partition-vel} a good transport field despite not being the conditional velocity in general. Empirically it reduces the burden on the corrector as shown in the experiment section.

\section{Experiments}
\label{sec:experiments}

We evaluate \method\ on three domain-transfer benchmarks that test logical composition under non-Gaussian endpoints.
In each setting we train a stochastic interpolant transporting a \emph{source} distribution $\rho_0$ to a \emph{target} distribution $\rho_1$, then sample from $\rho_1$ subject to a Boolean constraint $\varphi$ over semantic attributes.
Atoms are class-conditional velocity and score networks, plus noise-aware classifiers for the posterior weights in disjunctions and negations (\cref{sec:exp-details}). All results are averaged over 3 training and evaluation runs.

\textbf{Baselines and compute matching.}
We compare our method against two composition-only baselines without score re-equilibration ($K{=}0$):
\textbf{CompVel}, deterministic ODE integration of the composed proxy velocity in \cref{eq:partition-vel} (the analogue of LogDiff-style guidance \citep{alesiani2026logical}), and
\textbf{CompSI}, SDE integration that combines that velocity with the exactly composed score.
Our sampler is run as \textbf{\method-ODE} and \textbf{\method-SDE}, using the ODE or SDE predictor of \cref{alg:logsi} with $K{>}0$ Langevin correctors at fixed $t$.
For the logical \texttt{AND} we additionally report \textbf{Oracle-ODE/SDE}: compute-matched CFG on the joint label (guidance weight $1$), without composition or score correction.
All main-text comparisons are \emph{compute-matched} in number of function evaluations (NFE): each \method\ run with $N$ outer steps and $K$ correctors is matched against the corresponding $K{=}0$ baseline with enough extra integration steps to equalise the budget (\method-ODE against CompVel, \method-SDE against CompSI), which isolates score-based re-equilibration from simply spending more network calls on transport.

\textbf{Metrics.}
Our primary metric is logical \emph{conformity}: the fraction of generated samples whose attributes satisfy $\varphi$. On the image benchmarks attributes are read off by a held-out evaluator classifier, distinct from the noise-aware classifier that drives sampling. On QM9 they are recomputed from the generated structure with RDKit \citep{rdkit}, along with chemical validity, and conformity is measured only over valid molecules. For images we additionally report FID against an expression-conditioned reference set (real images whose ground-truth attributes satisfy the same $\varphi$) and LPIPS to the source image. For QM9 we report validity$\times$conformity, source-to-sample Kabsch root mean square deviation (RMSD), and novelty. Full definitions are in \cref{sec:exp-metrics}.

\begin{figure}[t]
    \centering
    \includegraphics[width=1.0\textwidth]{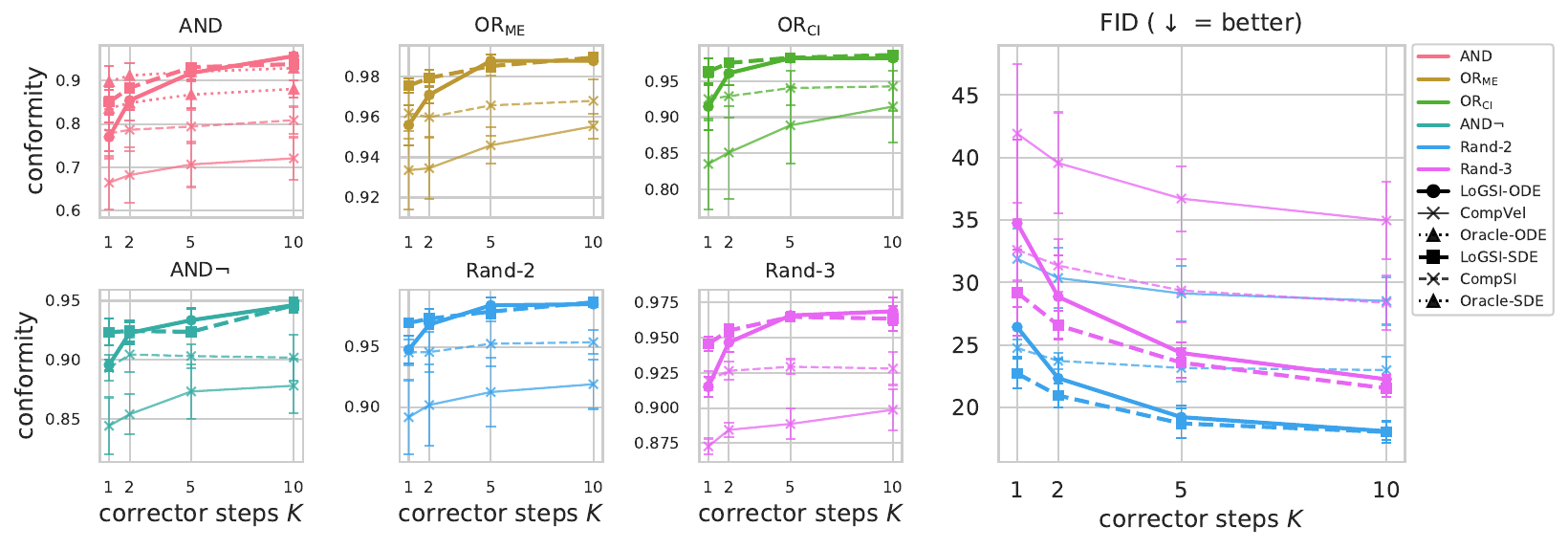}
    \caption{%
        \textbf{EMNIST to CMNIST} logical sampling at matched NFE.
        \textbf{(left)}~Conformity of \method\ and of the compute-matched baselines, as a function of corrector steps~$K$. Rand-2/Rand-3 are random depth-2/3 expressions.
        \textbf{(right)}~FID as a function of corrector steps~$K$ at matched NFE. For overview reasons only Rand-2/Rand-3 are shown, for all of the experiments see \cref{fig:additional_cmnist_results}.
        }
    \label{fig:emnist_results}
\end{figure}

\textbf{EMNIST to CMNIST transfer.}
\label{sec:exp-cemnist}
We train paired ResNet \citep{he:2016} velocity and score models transporting greyscale EMNIST \citep{cohen2017emnist} letters (\texttt{a--e}, \texttt{A--E}) to Colored MNIST digits \citep{lecun:1998, gaudi:2025}, so each source letter is edited into a digit satisfying the logical rule. Atoms are \texttt{Digit}$(d)$ and \texttt{Color}$(c)$ for $d,c\in\{0,\ldots,9\}$. We evaluate fixed templates (AND, OR\textsubscript{ME}, OR\textsubscript{CI}, AND$\neg$) and randomly generated expression trees of depth two and three over the same atoms. \cref{app:datasets} states which disjunction rule applies to which formula.
\Cref{fig:emnist_results} summarizes the compute-matched comparison. Adding correctors improves conformity by up to $14$\,pp over CompSI in the AND case and up to $4$\,pp on random depth-three expressions, where the smaller gain may be caused by the already high conformity $>96\%$. FID also decreases with more correctors, falling below the baselines, so re-equilibration improves fidelity as well as conformity rather than trading one for the other.
OR\textsubscript{ME} still benefits from correctors because \cref{thm:not-orme-exact} is exact only for the true fields, whereas learned velocities, classifier posteriors, and discretization remain approximate.
ODE and SDE transport are indistinguishable at larger $K$. Oracle-SDE matches the baselines while being stronger at small $K$, and Oracle-ODE is worse throughout.
Further metrics are in \cref{fig:additional_cmnist_results}: LPIPS to the source grows with $K$ but stays below the LPIPS of random source-target pairs, so edits remain close to their source.

\begin{figure}[t]
    \centering
    \includegraphics[width=1.0\textwidth]{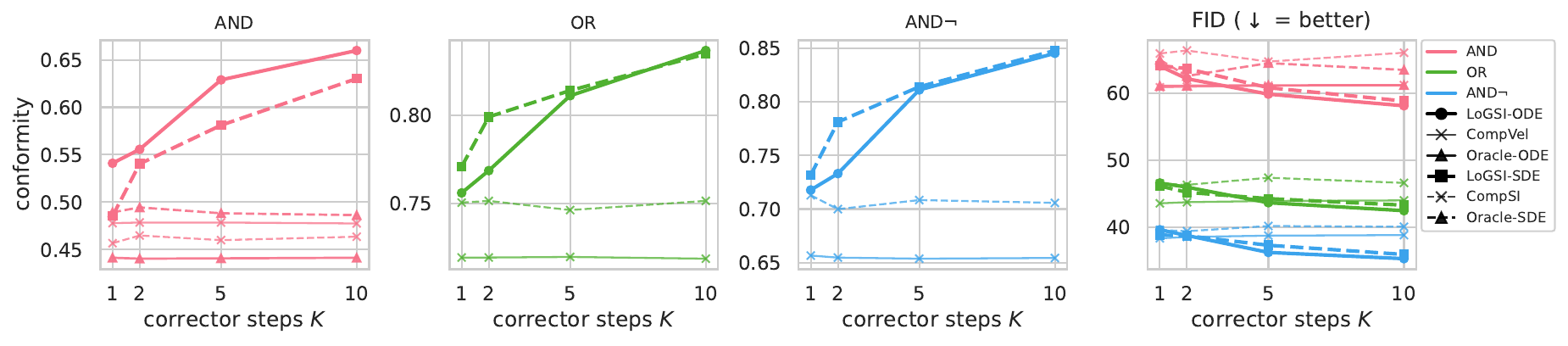}
    \caption{%
        \textbf{CelebA latent transfer} with logical constraints on \texttt{Hat} and \texttt{Glasses}.
        \textbf{(left)}~Conformity  of \method\ and the compute-matched baselines as a function of corrector steps~$K$.
        \textbf{(right)}~FID as a function of corrector steps~$K$ at matched NFE.
        Colours index the logical template (AND, OR, AND$\neg$). For readability, the plots with the error bars are moved to the appendix in \cref{fig:additional_celeba_results}. 
        }
    \label{fig:celeba_results}
\end{figure}

\begin{figure}[t]
    \centering
    \includegraphics[width=1.0\textwidth]{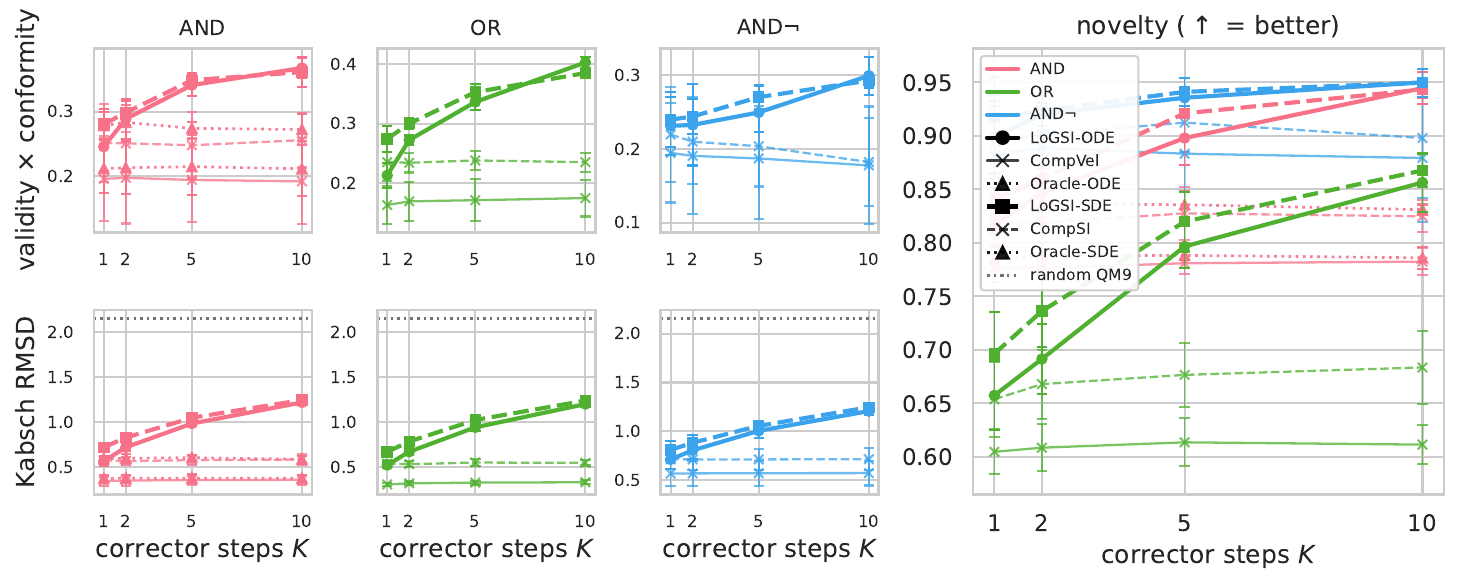}
    \caption{%
        \textbf{QM9 molecular editing} with logical constraints on \texttt{rings} and \texttt{oxygen}, at matched NFE.
        \textbf{(left, top)}~validity$\times$conformity and \textbf{(left, bottom)}~Kabsch-aligned RMSD to the source molecule, both as functions of corrector steps~$K$. The dotted line marks the RMSD between random QM9 molecules.
        \textbf{(right)}~Novelty of generated molecules vs.~$K$.
        }
    \label{fig:qm9_results}
\end{figure}

\textbf{CelebA transfer.}
\label{sec:exp-celeba}
Inspired by \cite{shi2023diffusion}, we transport CelebA \citep{liu2015deep} \emph{female} images ($\rho_0$) to \emph{male} images ($\rho_1$), training two SiT models \citep{ma2024sit} for velocity and score in the FLUX.1-schnell VAE \citep{flux_schnell_2024} latent space. Logical atoms are the two attributes \texttt{Hat} and \texttt{Glasses}, and we test the statements AND, OR\textsubscript{CI}, and AND$\neg$. Posterior weights for OR\textsubscript{CI} and negation come from a time-conditioned latent SiT classifier (\cref{sec:exp-details}).
\Cref{fig:celeba_results} reports the compute-matched results. Conformity again rises with correctors, by up to $18$\,pp over the baseline. Similarly to CMNIST, FID also decreases past the baseline values, and the ODE and SDE predictors perform comparably.
Additionally, both Oracle baselines perform worse.
As on CMNIST, LPIPS to the source image grows with $K$ but stays below the random source--target LPIPS (\cref{fig:additional_celeba_results}), so edited faces remain close to their source seed.

\textbf{QM9 molecular editing.}
\label{sec:exp-qm9}
On QM9 \citep{ramakrishnan2014quantum, ruddigkeit2012enumeration} we train EGNN \citep{satorras2021n} (as in \citet{hoogeboom2022equivariant}) velocity and score models that bridge \emph{source} to \emph{target} molecular graphs, both drawn from QM9. Unlike mirror interpolants that connect a sample to itself, we use a random coupling so that any molecule can be transported to any other, enabling editing. Atoms are binary molecular attributes. Our main statement pairs \texttt{has\_rings}$=1$ with \texttt{has\_oxygen}$=1$ under AND, OR\textsubscript{CI}, and AND$\neg$. Each sample is initialized from a concrete source graph whose labels \emph{violate} $\varphi$, so conformity measures whether the sampler can \emph{create} the constraint from a non-conforming start (\cref{sec:exp-details}).
\Cref{fig:qm9_results} reports the compute-matched results, a stricter test than the image benchmarks since a sample must be chemically valid \emph{and} rule-conforming. Validity$\times$conformity rises with the corrector budget by up to $16$\,pp over the baseline. Decomposing the product (\cref{fig:additional_qm9_results}) reveals a trade-off: conformity among valid molecules rises while validity loses. Since exact re-equilibration targets $p_t(\cdot \mid \varphi)$, whose samples are valid by construction, we read this loss as finite-$K$ approximation error rather than as a property of the guidance target.
As expected, the Kabsch RMSD to the source grows with $K$, reaching around $1.4\,\text{\AA}$, but stays well below the RMSD between random QM9 molecules (dotted line), so edits remain local. Novelty likewise grows with $K$. As before, ODE and SDE transport differ little at large $K$, and both Oracle baselines perform worse than the proposed method.

\begin{table}[t]
\caption{\textbf{Transport-velocity ablation.} Conformity (validity$\times$conformity for QM9) if the transport step uses the Gaussian-path proxy velocity (\method) vs.\ the unconditional velocity (uncd.), at $5$ corrector steps. Furthermore, we report also a baseline where no transport step was used. We report the AND case only, to avoid confounding from the noise-aware classifier used in OR/negation.}
\label{tab:transport_ab_ode}
\begin{center}
\begin{tabular}{lccc}
\toprule
\multicolumn{1}{c}{\bf Method}  &\multicolumn{1}{c}{\bf CMNIST}&\multicolumn{1}{c}{\bf CelebA}&\multicolumn{1}{c}{\bf QM9} 
\\ \midrule
\method\ 5 step        & $\mathbf{0.904 \pm 0.003}$ & $\mathbf{0.63 \pm 0.09}$ & $\mathbf{0.34 \pm 0.02}$\\
uncd. 5 step           & $0.150 \pm 0.004$ & $0.07 \pm 0.03$ & $0.27 \pm 0.02$\\
$\tfrac{1}{2}v_t(x|c_1)+\tfrac{1}{2}v_t(x|c_2)$ 5 step            & $0.650 \pm 0.007$ & $0.29 \pm 0.08$ & $0.31 \pm 0.02$\\
only score 5 step      & $0.034 \pm 0.001$ & $0.030 \pm 0.002$ & $0.053 \pm 0.007$\\
\bottomrule
\end{tabular}
\end{center}
\end{table}

\textbf{Transport-velocity ablation.}
It is a priori unclear which transport velocity $\tilde v_t$ is best, so we compare our Gaussian-path proxy velocity \cref{eq:partition-vel} against the unconditional velocity $\tilde v_t=v_t$, against no transport step at all, and against the equal-weight average $\tfrac{1}{2}v_t(x\mid c_1)+\tfrac{1}{2}v_t(x\mid c_2)$, i.e.\ \citet{liu2022compositional}'s weighted score composition applied to velocities instead of scores (\cref{tab:transport_ab_ode,tab:transport_ab_ode2}). The proxy velocity yields higher conformity across all three benchmarks, confirming the design choice of \cref{sec:proxy-vel}. We report ODE transport here, and the SDE variant (\cref{tab:transport_ab_sde}) shows the same trend.

\section{Conclusion}
We studied logical composition for stochastic interpolants that transport between non-Gaussian source and target distributions.
While conditional scores compose exactly in the LogDiff sense on general interpolants, the analogous velocity rule survives only in part: negation and exclusive disjunction remain exact on any interpolant, whereas conjunction, and hence any formula containing one, is exact only on Gaussian-path bridges.
\method\ therefore separates roles: a proxy composed velocity advances samples along the interpolant, and exactly composed conditional scores re-equilibrate them toward $p_t(\cdot\mid\varphi)$ at each timestep.
Across EMNIST-to-CMNIST, CelebA, and QM9 editing benchmarks, this transport--re-equilibration procedure improves logical conformity at matched network-evaluation budget over composed-velocity baselines, for both fixed Boolean templates and randomly generated formulas. It additionally improves fidelity (FID) on the image benchmarks and novelty on QM9.
More broadly, the same score-based viewpoint may extend to other bridge learners and to settings where logical combinations of constraints arise together with non-Gaussian transport.

\bibliography{bib/FKC,bib/guidance,bib/references,bib/ab_pref,bib/alignab,bib/si}
\bibliographystyle{arxiv2027_conference}

\clearpage
\appendix

\section{Method Details}

\subsection{Recursive composition rules}

\paragraph{Scores.}
\label{app:recursive_scores}
As stated in the main text, exact composition is possible whenever $\varphi$ admits a probabilistic-circuit representation whose conjunction nodes combine conditionally independent subformulas and whose disjunction nodes combine either conditionally independent or mutually exclusive subformulas \citep{alesiani2026logical}. 
In that case, $p_t(\varphi\mid x)$ and $s_t(\varphi\mid x)$ can be computed recursively from the atomic quantities. Writing $\widehat\pi(\psi)= p_t(\psi\mid x)$ for posterior estimates of subformulas $\psi$, the main rules are
\begin{align}
        s_t(c_i\mid x) &= \nabla_x \log p_t(x\mid c_i)-\nabla_x \log p_t(x),&\text{(Atomic)}\\
        s_t(\lnot\psi\mid x) &= -\frac{\widehat\pi(\psi)}{1-\widehat\pi(\psi)}\,s_t(\psi\mid x),&\text{(NOT)}\\
        s_t(\psi\land\chi\mid x) &= s_t(\psi\mid x)+s_t(\chi\mid x),&\text{(AND)}\\
        s_t(\psi\lor\chi\mid x) &= \frac{\widehat\pi(\psi)\widehat\pi(\lnot\chi)\,s_t(\psi\mid x)+\widehat\pi(\chi)\widehat\pi(\lnot\psi)\,s_t(\chi\mid x)}{\widehat\pi(\psi)+\widehat\pi(\chi)-\widehat\pi(\psi)\widehat\pi(\chi)}
        ,\quad&\text{(OR-CI)}\\
        s_t(\psi\lor\chi\mid x) &= \frac{\widehat\pi(\psi)\,s_t(\psi\mid x)+\widehat\pi(\chi)\,s_t(\chi\mid x)}{\widehat\pi(\psi)+\widehat\pi(\chi)},
        \quad&\text{(OR-ME)}
\end{align}
with matching recursive rules for $\widehat\pi(\varphi)$ given in \cite{alesiani2026logical}. For each disjunction one uses the OR-CI rule if the subformulas are conditionally independent given $x_t=x$, and the OR-ME rule if they are mutually exclusive.

\paragraph{Velocities.}
\label{app:recursive_vel}
Under the same conditions as above, the velocities of Gaussian path \ac{fm} combine in a very similar way as the scores. Here $v(x\mid c_i)$ is the conditional velocity field and $v(x)$ is the unconditional velocity field. By \cref{thm:not-orme-exact} the Atomic, NOT and OR-ME rules below are exact for velocities on any stochastic interpolant. Only the AND rule, and through inclusion--exclusion the OR-CI rule, require the Gaussian-path assumption, with the error of the AND rule given in closed form by \cref{thm:and-bias}.
\begin{align}
        v_t(c_i\mid x) &= v_t(x\mid c_i)-v_t(x),&\text{(Atomic)}\\
        v_t(\lnot\psi\mid x) &= -\frac{\widehat\pi(\psi)}{1-\widehat\pi(\psi)}\,v_t(\psi\mid x),&\text{(NOT)}\\
        v_t(\psi\land\chi\mid x) &= v_t(\psi\mid x)+v_t(\chi\mid x),&\text{(AND)}\\
        v_t(\psi\lor\chi\mid x) &= \frac{\widehat\pi(\psi)\widehat\pi(\lnot\chi)\,v_t(\psi\mid x)+\widehat\pi(\chi)\widehat\pi(\lnot\psi)\,v_t(\chi\mid x)}{\widehat\pi(\psi)+\widehat\pi(\chi)-\widehat\pi(\psi)\widehat\pi(\chi)}
        ,\quad&\text{(OR-CI)}\\
        v_t(\psi\lor\chi\mid x) &= \frac{\widehat\pi(\psi)\,v_t(\psi\mid x)+\widehat\pi(\chi)\,v_t(\chi\mid x)}{\widehat\pi(\psi)+\widehat\pi(\chi)}.
        \quad&\text{(OR-ME)}
\end{align}
The recursive rules for the noise aware classifiers $\widehat\pi(\varphi)$ stay also the same as above.

\subsection{Proof of Gaussian-path results}
\label{sec:proof-gp-fm}

\begin{lemma}[Affine velocity--score relation]
\label{lem:affine-vs}
Let $x_t=\alpha(t)x_1+\beta(t)x_0$ with $x_0\sim\mathcal N(0,I)$ independent of $x_1$, and let $c$ be a terminal condition determined by $x_1$ whose noising kernel is independent of $c$ given $x_1$. Then for every $t\in(0,1)$
\begin{align}
        v_t(x\mid c)=a_t x+b_t s_t(x\mid c),\quad\text{ with }\quad
        a_t=\frac{\dot\alpha(t)}{\alpha(t)},\quad
        b_t=\frac{\dot\alpha(t)}{\alpha(t)}\beta^2_t-\dot\beta(t)\beta(t),
        \label{eq:affine-vel-score}
\end{align}
where the coefficients $(a_t,b_t)$ depend on $t$ only, not on $c$. In particular $v_t(x)=a_t x+b_t s_t(x)$.
\end{lemma}

\begin{proof}[Proof of \cref{lem:affine-vs}]
Let $x_t=\alpha(t)x_1+\beta(t)x_0$ with $x_0\sim\mathcal N(0,I)$, and let $c$ be any conditioning event under which the noising kernel does not depend on $c$ directly, only through the law of $x_1$:
\begin{align}\label{eq:noising_kernel}
p_t(x\mid x_1,c) = p_t(x\mid x_1) = \mathcal N\!\big(x;\,\alpha(t)x_1,\,\beta(t)^2 I\big).
\end{align}
Since $x_0=(x_t-\alpha(t)x_1)/\beta(t)$ holds surely, taking the conditional expectation defining the marginal velocity field,
\begin{align}\label{eq:marginal_velocity}
v_t(x\mid c) = \mathbb E\big[\dot\alpha(t)x_1+\dot\beta(t)x_0 \,\big|\, x_t=x,\,c\big]
= \frac{\dot\beta(t)}{\beta(t)}\,x \;+\; \left(\dot\alpha(t)-\frac{\dot\beta(t)}{\beta(t)}\alpha(t)\right)\mathbb E[x_1\mid x_t=x,c],    
\end{align}

using that $x_1$ enters linearly, so the substitution can be pushed through the expectation term by term.

By \eqref{eq:noising_kernel} and Bayes' rule, $p(x_1\mid x_t=x,c)\propto p_t(x\mid x_1)\,q(x_1\mid c)$, and differentiating $p_t(x\mid c)=\int p_t(x\mid x_1)q(x_1\mid c)\,dx_1$ under the integral sign with $\nabla_x\log p_t(x\mid x_1) = -(x-\alpha(t)x_1)/\beta(t)^2$ gives the Tweedie-type identity
\begin{align}\label{eq:tweedie}
\mathbb E[X_1\mid X_t=x,c] = \frac{x+\beta(t)^2\,s_t(x\mid c)}{\alpha(t)}, \qquad s_t(x\mid c):=\nabla_x\log p_t(x\mid c).
\end{align}

Substituting \eqref{eq:tweedie} into \eqref{eq:marginal_velocity}:
\begin{align}
v_t(x\mid c) &= \frac{\dot\beta(t)}{\beta(t)}\,x + \left(\dot\alpha(t)-\frac{\dot\beta(t)}{\beta(t)}\alpha(t)\right)\frac{x+\beta(t)^2 s_t(x\mid c)}{\alpha(t)}
\\&= \frac{\dot\alpha(t)}{\alpha(t)}\,x + \left(\frac{\dot\alpha(t)}{\alpha(t)}\beta(t)^2-\dot\beta(t)\beta(t)\right) s_t(x\mid c).        
\end{align}
Thus $v_t(x\mid c)=a_t x+b_t s_t(x\mid c)$ with $a_t,b_t$ as in the statement, independent of $c$. Taking $c$ to be the trivial conditioning event gives $v_t(x)=a_t x+b_t s_t(x)$.
\end{proof}

\begin{proof}[Proof of \cref{prop:vel-compose}]
By \cref{lem:affine-vs}, for any conditioning label $c$,
\begin{align}
        v_t(x\mid c)=a_t x+b_t s_t(x\mid c).
\end{align}
Hence
\begin{align}
        v_t(x\mid\varphi)
        &=a_t x+b_t\nabla_x\log p_t(x\mid\varphi)\notag\\
        &=a_t x+b_t\left(s_t(x)+\sum_i\alpha_i^{\varphi}(t,x)\bigl(s_t(x\mid c_i)-s_t(x)\bigr)\right)\notag\\
        &=a_t x+b_t s_t(x)+\sum_i\alpha_i^{\varphi}(t,x)\,b_t\bigl(s_t(x\mid c_i)-s_t(x)\bigr)\notag\\
        &=v_t(x)+\sum_i\alpha_i^{\varphi}(t,x)\bigl(v_t(x\mid c_i)-v_t(x)\bigr),
\end{align}
because we used \eqref{eq:logdiff_lin_comb_score_xphi} in the first step, and the $a_tx$ terms cancel identically in the last substitution.
\end{proof}

\subsection{Logical rules for velocities}
\label{sec:proof-vel-obstruction}

We first record the reduction of the problem to the drift, then state and prove \cref{lem:moment-algebra}, prove \cref{thm:not-orme-exact} including its circuit consequence, prove \cref{thm:and-bias} in both of its forms, and finally give the explicit counterexample.

\paragraph{Obstruction in the drift.}
Using \eqref{eq:general-vel-decomp}, expand $\check v_t$ from \eqref{eq:partition-vel} as
\begin{align}
        \check v_t(x\mid\varphi)
        &=\check\mu_t(x\mid\varphi)-\dot\gamma(t)\gamma(t)\,s_t(x\mid\varphi),
        \label{eq:app-check-v-decomp}
\end{align}
where $\check\mu_t(x\mid\varphi):=\mu_t(x)+\sum_i\alpha_i^{\varphi}(t,x)\bigl(\mu_t(x\mid c_i)-\mu_t(x)\bigr)$ and we used the exact score identity \eqref{eq:logdiff_lin_comb_score_xphi}. Comparing with $v_t(x\mid\varphi)=\mu_t(x\mid\varphi)-\dot\gamma(t)\gamma(t)\,s_t(x\mid\varphi)$, the score terms cancel and
\begin{align}
        \check v_t(x\mid\varphi)-v_t(x\mid\varphi)=\check\mu_t(x\mid\varphi)-\mu_t(x\mid\varphi).
        \label{eq:app-obstruction}
\end{align}
Thus $\check v_t=v_t(\cdot\mid\varphi)$ iff
\begin{align}
        \mu_t(x\mid\varphi)=\mu_t(x)+\sum_i\alpha_i^{\varphi}(t,x)\bigl(\mu_t(x\mid c_i)-\mu_t(x)\bigr).
        \label{eq:app-mu-obstruction}
\end{align}
Therefore, the property of velocity compositionality reduces to the property of the compositionality of drifts.

\paragraph{Drift moments and Boolean algebra.}

\begin{lemma}[Drift moments and Boolean algebra]
\label{lem:moment-algebra}
Let $t\in(0,1)$ and $x$ with $p_t(x)>0$, and let $\psi,\chi$ be subformulas with $p_t(\psi\mid x)=\widehat\pi(\psi)>0$. Then, with $m^{\psi}_t$ as in \cref{eq:drift-moment},
\begin{align}
        \mu_t(x\mid\psi)=\frac{m^{\psi}_t(x)}{\widehat\pi(\psi)},\quad
        m^{\lnot\psi}_t(x)=\mu_t(x)-m^{\psi}_t(x),\quad
        m^{\psi\lor\chi}_t(x)=m^{\psi}_t(x)+m^{\chi}_t(x)-m^{\psi\land\chi}_t(x),
        \label{eq:moment-algebra}
\end{align}
together with $\widehat\pi(\lnot\psi)=1-\widehat\pi(\psi)$ and $\widehat\pi(\psi\lor\chi)=\widehat\pi(\psi)+\widehat\pi(\chi)-\widehat\pi(\psi\land\chi)$. These identities hold for any interpolant $I$, noise schedule $\gamma$, coupling $\pi$ and endpoints $\rho_0,\rho_1$, with no independence or exclusivity assumption.
\end{lemma}

\begin{proof}[Proof of \cref{lem:moment-algebra}]
Fix $t\in(0,1)$ and $x$ with $p_t(x)>0$, and let $\pi_t(\cdot\mid x)$ be the conditional density of $(x_0,x_1)$ given $x_t=x$, so that $m^{\psi}_t(x)=\int\partial_t I\,\mathbf 1_\psi\,d\pi_t(\cdot\mid x)$ and $\widehat\pi(\psi)=\int\mathbf 1_\psi\,d\pi_t(\cdot\mid x)$. For $\widehat\pi(\psi)>0$, the conditional law of $(x_0,x_1)$ given $\{x_t=x\}\cap\{x_1\models\psi\}$ is $\pi_t(\cdot\mid x)$ restricted to $\{x_1\models\psi\}$ and renormalised by $\widehat\pi(\psi)$, hence

\begin{align}
        \mu_t(x\mid\psi)=\frac{\int\partial_t I\,\mathbf 1_\psi\,d\pi_t(\cdot\mid x)}{\widehat\pi(\psi)}=\frac{m^{\psi}_t(x)}{\widehat\pi(\psi)}.
\end{align}
Both $\psi\mapsto m^{\psi}_t(x)$ and $\psi\mapsto\widehat\pi(\psi)$ are linear in the indicator $\mathbf 1_\psi$, so the remaining identities follow from the pointwise Boolean identities
\begin{align}
        \mathbf 1_{\lnot\psi}=1-\mathbf 1_{\psi},\qquad
        \mathbf 1_{\psi\land\chi}=\mathbf 1_{\psi}\mathbf 1_{\chi},\qquad
        \mathbf 1_{\psi\lor\chi}=\mathbf 1_{\psi}+\mathbf 1_{\chi}-\mathbf 1_{\psi}\mathbf 1_{\chi}.
        \label{eq:app-bool-indicators}
\end{align}
\end{proof}

\paragraph{Exact rules: negation and exclusive disjunction.}

\begin{proof}[Proof of \cref{thm:not-orme-exact}]
Writing $v_t(\psi\mid x):=v_t(x\mid\psi)-v_t(x)$ for the logical velocity of a subformula, the two rules to be proved are
\begin{align}
        v_t(\lnot\psi\mid x)&=-\frac{\widehat\pi(\psi)}{1-\widehat\pi(\psi)}\,v_t(\psi\mid x)
        &&\text{if }\widehat\pi(\psi)\in(0,1),
        \label{eq:app-not-rule}\\
        v_t(\psi\lor\chi\mid x)&=\frac{\widehat\pi(\psi)\,v_t(\psi\mid x)+\widehat\pi(\chi)\,v_t(\chi\mid x)}{\widehat\pi(\psi)+\widehat\pi(\chi)}
        &&\text{if }\mathbf 1_\psi\mathbf 1_\chi\equiv0,\ \widehat\pi(\psi)+\widehat\pi(\chi)>0,
        \label{eq:app-orme-rule}
\end{align}
which are the NOT and OR-ME rules of \cref{app:recursive_vel}. Write $q:=\widehat\pi(\psi)$, $r:=\widehat\pi(\chi)$ and $b_t:=-\dot\gamma(t)\gamma(t)$, so that \cref{eq:general-vel-decomp} reads $v_t(\cdot\mid c)=\mu_t(\cdot\mid c)+b_t\,s_t(\cdot\mid c)$ with $b_t$ independent of the conditioning event. It suffices to show that the linear combinations of the drifts is the same as the scores (\cref{app:recursive_scores}).

\emph{Negation.} By \cref{lem:moment-algebra}, $m^{\lnot\psi}_t(x)=\mu_t(x)-q\,\mu_t(x\mid\psi)$ and $\widehat\pi(\lnot\psi)=1-q$, so
\begin{align}
        \mu_t(x\mid\lnot\psi)-\mu_t(x)
        =\frac{\mu_t(x)-q\,\mu_t(x\mid\psi)}{1-q}-\mu_t(x)
        =-\frac{q}{1-q}\bigl[\mu_t(x\mid\psi)-\mu_t(x)\bigr].
\end{align}
The coefficients agree with the rules for the score (\cref{app:recursive_scores}), which gives the NOT rule.

\emph{Exclusive disjunction.} If $\mathbf 1_{\psi}\mathbf 1_{\chi}\equiv0$ then $m^{\psi\lor\chi}_t=m^{\psi}_t+m^{\chi}_t$ and $\widehat\pi(\psi\lor\chi)=q+r$ by \cref{lem:moment-algebra} and \cref{eq:app-bool-indicators}, hence
\begin{align}
        \mu_t(x\mid\psi\lor\chi)-\mu_t(x)
        =\frac{q\bigl[\mu_t(x\mid\psi)-\mu_t(x)\bigr]+r\bigl[\mu_t(x\mid\chi)-\mu_t(x)\bigr]}{q+r}.
\end{align}
Here, the weights are again the same as the scores (\cref{app:recursive_scores}). Adding $b_t$ times the score identities to the drift identities yields \cref{eq:app-not-rule,eq:app-orme-rule}.

\emph{Conjunction-free circuits.} The second claim follows by induction over the circuit. At a leaf, $v_t(c_i\mid x)=v_t(x\mid c_i)-v_t(x)$ holds by definition and $\widehat\pi(c_i)$ in practice is supplied by the noise-aware classifier. If the children of a node are exact, then \cref{eq:app-not-rule,eq:app-orme-rule} show that a negation or exclusive-disjunction node returns the exact logical velocity of its subformula, and the accompanying posterior recursions $\widehat\pi(\lnot\psi)=1-\widehat\pi(\psi)$ and $\widehat\pi(\psi\lor\chi)=\widehat\pi(\psi)+\widehat\pi(\chi)$ are exact by \cref{lem:moment-algebra}. The value at the root is therefore $v_t(x\mid\varphi)-v_t(x)$, i.e.\ $\check v_t(\cdot\mid\varphi)=v_t(\cdot\mid\varphi)$. A conditionally independent disjunction is not covered, because $\mathbf 1_{\psi\lor\chi}=\mathbf 1_{\psi}+\mathbf 1_{\chi}-\mathbf 1_{\psi}\mathbf 1_{\chi}$ contains a conjunction moment. This obstruction, which this creates is covered in \cref{thm:and-bias}.
\end{proof}

\paragraph{The obstruction at conjunction.}

\begin{proof}[Proof of \cref{thm:and-bias}]
Define the following function,
\begin{align}
        R_t(x):=\mathbb E\Bigl[\partial_t I(t,x_0,x_1)\bigl(\mathbf 1_{c_1}-\widehat\pi(c_1)\bigr)\bigl(\mathbf 1_{c_2}-\widehat\pi(c_2)\bigr)\Bigm| x_t=x\Bigr].
        \label{eq:residual-moment}
\end{align}
We first show that the difference of proxy and true velocity equals $-R_t(x)/\bigl(\widehat\pi(c_1)\widehat\pi(c_2)\bigr)$, and then convert this into the form given in the main text \cref{eq:and-bias-cell}.

For $S\subseteq\{1,2\}$ put $M_S:=m^{\bigwedge_{i\in S}c_i}_t(x)$ and $Q_S:=\widehat\pi(\bigwedge_{i\in S}c_i)$, with $M_\emptyset=\mu_t(x)$ and $Q_\emptyset=1$, and abbreviate $M_i:=M_{\{i\}}$, $Q_i:=Q_{\{i\}}$, $M_{12}:=M_{\{1,2\}}$, $Q_{12}:=Q_{\{1,2\}}$; by \cref{lem:moment-algebra}, $M_i=Q_i\,\mu_t(x\mid c_i)$. Expanding the product in \cref{eq:residual-moment} and using linearity of the conditional expectation,
\begin{align}
        R_t(x)=M_{12}-Q_2M_1-Q_1M_2+Q_1Q_2M_\emptyset .
        \label{eq:app-residual-expanded}
\end{align}
Conditional independence gives $Q_{12}=Q_1Q_2>0$, so dividing \cref{eq:app-residual-expanded} by $Q_1Q_2$ and using $\mu_t(x\mid\varphi)=M_{12}/Q_{12}$ from \cref{lem:moment-algebra},
\begin{align}
        \mu_t(x\mid\varphi)
        =\frac{R_t(x)}{Q_1Q_2}+\frac{M_1}{Q_1}+\frac{M_2}{Q_2}-M_\emptyset
        =\frac{R_t(x)}{Q_1Q_2}+\check\mu_t(x\mid\varphi),
\end{align}
because the AND coefficients are $\alpha^{\varphi}_1=\alpha^{\varphi}_2=1$, for which $\check\mu_t(x\mid\varphi)=\mu_t(x\mid c_1)+\mu_t(x\mid c_2)-\mu_t(x)$. Conditional independence also makes the composed score exact, so \cref{eq:app-obstruction} applies and
\begin{align}
        \check v_t(x\mid\varphi)-v_t(x\mid\varphi)
        =\check\mu_t(x\mid\varphi)-\mu_t(x\mid\varphi)
        =-\frac{R_t(x)}{Q_1Q_2}.
        \label{eq:app-and-bias-moment}
\end{align}
Next we show how to convert this into the statement in the main text.

Write $A_i:=\mathbf 1_{c_i}$ and $p_i:=\widehat\pi(c_i)\in(0,1)$, and let all expectations be under $\pi_t(\cdot\mid x)$. All four cells then carry positive mass, so the cell-conditional drifts $d_t(a_1,a_2)$ are well defined. Conditioning additionally on the pair $(A_1,A_2)$ and using the tower property,
\begin{align}
        R_t(x)=\mathbb E\bigl[d_t(A_1,A_2)\,(A_1-p_1)(A_2-p_2)\bigr].
\end{align}
Since $(a_1,a_2)\mapsto d_t(a_1,a_2)$ is a function on $\{0,1\}^2$, it has the exact multilinear expansion
\begin{align}
        d_t(a_1,a_2)=\theta_\emptyset+\theta_1a_1+\theta_2a_2+\theta_{12}a_1a_2,
        \qquad \theta_{12}=\Delta_t(x),
        \label{eq:app-mobius-2}
\end{align}
with $\theta_\emptyset=d_t(0,0)$, $\theta_1=d_t(1,0)-d_t(0,0)$ and $\theta_2=d_t(0,1)-d_t(0,0)$; the coefficient of the product term is the mixed second difference \cref{eq:delta-def}. Conditional independence of $A_1,A_2$ together with $A_i^2=A_i$ gives
\begin{align}
        \mathbb E[(A_1-p_1)(A_2-p_2)]&=0,\qquad
        \mathbb E[A_i(A_1-p_1)(A_2-p_2)]=0\quad(i=1,2),\\
        \mathbb E[A_1A_2(A_1-p_1)(A_2-p_2)]&=\mathbb E[A_1(A_1-p_1)]\,\mathbb E[A_2(A_2-p_2)]=p_1(1-p_1)\,p_2(1-p_2),
\end{align}
so only the interaction term of \cref{eq:app-mobius-2} survives,
\begin{align}
        R_t(x)=p_1(1-p_1)\,p_2(1-p_2)\,\Delta_t(x).
        \label{eq:app-residual-interaction}
\end{align}
Substituting into \cref{eq:app-and-bias-moment} gives $\check v_t-v_t(\cdot\mid\varphi)=-R_t(x)/(p_1p_2)=-(1-p_1)(1-p_2)\Delta_t(x)$, which is \cref{eq:and-bias-cell}. Since $p_1p_2>0$, the bias vanishes if and only if $R_t(x)=0$, equivalently $(1-p_1)(1-p_2)\Delta_t(x)=0$, and the two limiting cases in the statement follow by inspection of the prefactor.
\end{proof}

\begin{remark}[Gaussian paths force the residual to vanish]
\label{rem:gp-vanishing}
On the Gaussian path of \cref{sec:gaussian-path}, $x_t=\alpha(t)x_1+\beta(t)x_0$ with $x_0\sim\mathcal N(0,I)$ independent of $x_1$ and $\gamma\equiv0$, the relation $x_0=(x_t-\alpha(t)x_1)/\beta(t)$ holds surely, so
\begin{align}
        \partial_t I(t,x_0,x_1)=\frac{\dot\beta(t)}{\beta(t)}\,x_t+\Bigl(\dot\alpha(t)-\frac{\dot\beta(t)}{\beta(t)}\alpha(t)\Bigr)x_1 .
\end{align}
Given $x_t=x$ the drift is thus an affine function of $x_1$ with coefficients that do not depend on the conditioning, and by the Tweedie identity \eqref{eq:tweedie} the conditional mean $\mathbb E[x_1\mid x_t=x,c]$ is an affine function of $s_t(x\mid c)$ with $c$-independent coefficients. Exact score composition \eqref{eq:logdiff_lin_comb_score_xphi} therefore propagates to the drifts, $\check\mu_t(x\mid\varphi)=\mu_t(x\mid\varphi)$, i.e.\ $R_t\equiv0$ in \cref{thm:and-bias}. This is the mechanism behind \cref{prop:vel-compose}. Since the two indicator variances are strictly positive for $\widehat\pi(c_i)\in(0,1)$, \cref{eq:app-residual-interaction} then also gives $\Delta_t\equiv0$: on a Gaussian path, conditional independence of the atomic \emph{events} already forces additivity of the cell-conditional \emph{drifts}. It is exactly this implication that non-Gaussian endpoints destroy.
\end{remark}

Now that we have seen that the problem is inside the drift part of the velocity field, we can construct an explicit FM counterexample.

\begin{figure}[t]
    \centering
    \includegraphics[width=1.0\textwidth]{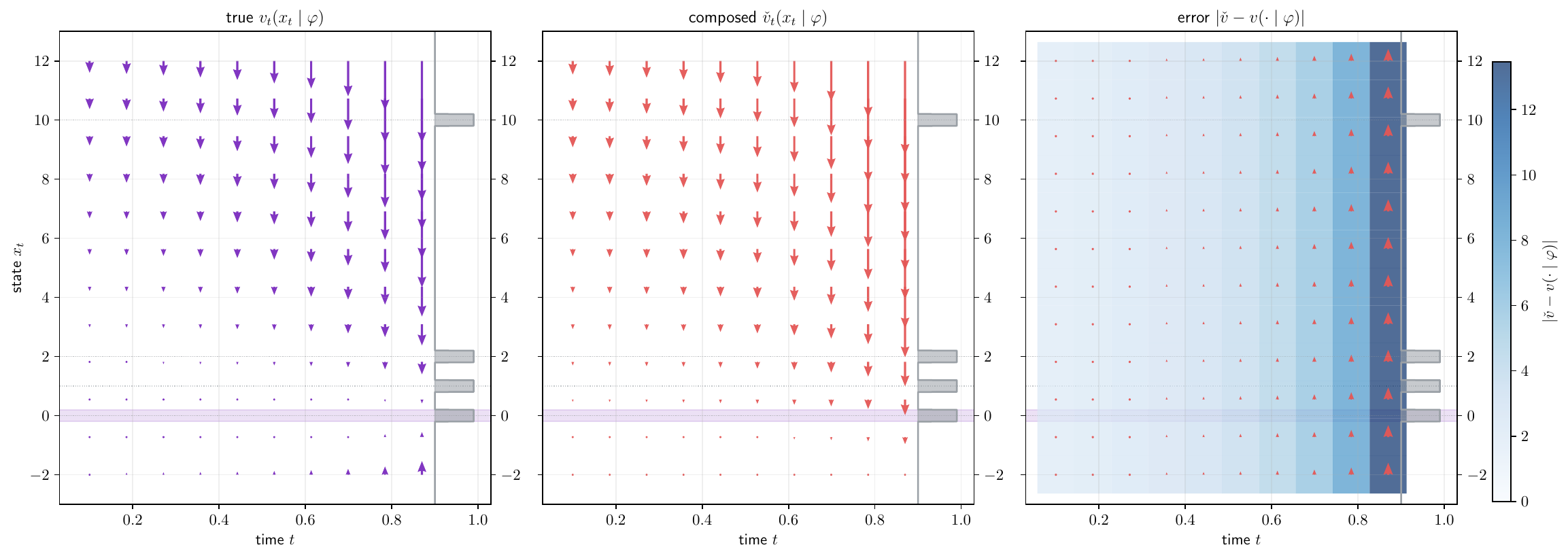}
    \caption{%
        \textbf{Explicit counterexample.} Here we plot the situation depicted in the counterexample Appendix \ref{par:explicit-counterexample}. On the left of each plot the source distribution and on the right the target distribution is shown. The first plot depicts the velocity field at different times of the AND composition. (Since the problem is one dimensional the velocity field is also one dimensional and has no component across time) The second plot depicts the linear combined atomic velocity fields. And the right shows the difference between these two, proving that velocities can not be combined together in the same way scores can. The depicted arrows of the velocity are downscaled by a factor $0.03$ for readability.
    }
    \label{fig:counterexample}
\end{figure}

\begin{proof}[\normalfont\textbf{Explicit counterexample.}]
\label{par:explicit-counterexample}
We now give a construction where the atoms are \emph{exactly} conditionally independent given $x_t$, so LogDiff's CI-AND score rule with $\alpha_1^{\varphi}=\alpha_2^{\varphi}=1$ is the exact composed score, while the LogDiff velocity composition still fails.

We use the flow-matching bridge ($\gamma\equiv 0$, $I(t,x_0,x_1)=(1-t)x_0+tx_1$) with independent coupling. Fix cluster centres $v_1=0,\,v_2=1,\,v_3=2,\,v_4=10$ and a small cluster half-width $\ell/2$ with $\ell\in(0,1)$ (so the intervals $B_j=[v_j-\ell/2,v_j+\ell/2]$ are disjoint). Take
\begin{align}
        \rho_1(x_1)=\tfrac{1}{4\ell}\sum_{j=1}^{4}\mathbf 1_{B_j}(x_1),\qquad
        \rho_0(x_0)=\tfrac{1}{2R}\mathbf 1_{[-R,R]}(x_0),
\end{align}
with $R$ large (concretely $R=100$). The source is non-Gaussian, so we are not on a Gaussian-path bridge. Define atoms and formula
\begin{align}
        c_1=\{x_1\in B_1\cup B_2\},\qquad
        c_2=\{x_1\in B_1\cup B_3\},\qquad
        \varphi=c_1\wedge c_2=\{x_1\in B_1\},
\end{align}
and evaluate at $(t,x_t)=(\tfrac12,5)$.
At $t=\tfrac12$, $x_t=(x_0+x_1)/2$, so a change of variables gives the conditional density of $x_t$ given $x_1$:
\begin{align}
        f_{x_t\mid x_1}(x_t\mid x_1)=2\,\rho_0(2x_t-x_1)=\tfrac{1}{R}\,\mathbf 1_{[-R,R]}(2x_t-x_1).
\end{align}
Define
\begin{align}
J:=\left(
\frac{-R+10+\ell/2}{2},
\frac{R-\ell/2}{2}
\right).
\end{align}
For every $x_t\in J$ and every $x_1\in\bigcup_j B_j\subset[-\ell/2,10+\ell/2]$, one has $2x_t-x_1\in[-R,R]$, so the indicator is $1$ and $f_{x_t\mid x_1}(x_t\mid x_1)=1/R$ is constant in $x_1$. By Bayes' rule,
\begin{align}
f_{x_1\mid x_t}(x_1\mid x_t)\;\propto\;\rho_1(x_1)\,f_{x_t\mid x_1}(x_t\mid x_1)\;\propto\;\rho_1(x_1),\qquad x_t\in J,
\end{align}
so the posterior of $x_1$ given $x_t$ equals $\rho_1$ throughout $J$, with mass $\tfrac14$ on each cluster $B_j$. In particular, $5\in J$, and throughout this neighborhood
\begin{align}
P(c_1\mid x_t)=\tfrac12,\qquad
P(c_2\mid x_t)=\tfrac12,\qquad
P(c_1\land c_2\mid x_t)=\tfrac14.
\end{align}
Thus $c_1\perp c_2\mid x_t$ holds locally around $x_t=5$.
The corresponding probability identity can therefore be differentiated
with respect to $x_t$, and LogDiff's CI-AND rule gives the exact composed
score at $x_t=5$.

\emph{Velocities and LogDiff velocity composition.} For $\gamma\equiv 0$ and linear $I$,
\begin{align}
        v_t(x_t\mid c)=\mathbb E[x_1-x_0\mid x_t,c]=\tfrac{1}{1-t}\bigl(\mathbb E[x_1\mid x_t,c]-x_t\bigr),
\end{align}
which at $t=\tfrac12$, $x_t=5$ becomes $2(\mathbb E[x_1\mid x_t,c]-5)$. Since each cluster $B_j$ is symmetric around its centre $v_j$ and receives uniform weight, $\mathbb E[x_1\mid x_t=5,\,x_1\in B_j]=v_j$ exactly, hence:
\begin{align}
\begin{array}{ll}
\mathbb E[x_1\mid x_t,c_1]=\tfrac{v_1+v_2}{2}=0.5, & v_t(5\mid c_1)=-9,\\[2pt]
\mathbb E[x_1\mid x_t,c_2]=\tfrac{v_1+v_3}{2}=1, & v_t(5\mid c_2)=-8,\\[2pt]
\mathbb E[x_1\mid x_t]=\tfrac{v_1+v_2+v_3+v_4}{4}=3.25, & v_t(5)=-3.5,\\[2pt]
\mathbb E[x_1\mid x_t,\varphi]=v_1=0, & v_t(5\mid\varphi)=-10.
\end{array}
\end{align}
The LogDiff velocity composition (with the correct CI-AND coefficients $\alpha_1^{\varphi}=\alpha_2^{\varphi}=1$) evaluates to
\begin{align}
        \check v_t(5\mid\varphi)&=v_t(5)+[v_t(5\mid c_1)-v_t(5)]+[v_t(5\mid c_2)-v_t(5)]\notag\\
        &=-3.5+(-5.5)+(-4.5)=-13.5\;\neq\;-10=v_t(5\mid\varphi).
\end{align}

\emph{Interpretation.} Consistent with \eqref{eq:app-obstruction}, $\check v_t(5\mid\varphi)-v_t(5\mid\varphi)=\check\mu_t(5\mid\varphi)-\mu_t(5\mid\varphi)$, and the residual is
\begin{align}
        \check\mu_t(5\mid\varphi)-\mu_t(5\mid\varphi)
        =\tfrac{v_2+v_3-v_1-v_4}{2}
        =\tfrac{1+2-0-10}{2}=-3.5,
\end{align}
matching $-13.5-(-10)=-3.5$. The drift residual vanishes exactly when the four values form a parallelogram, $v_2+v_3=v_1+v_4$, a condition independent of the CI condition on the atomic probabilities. Choosing $\{v_j\}$ to satisfy uniform posterior (which yields CI at $x_t$) but to violate $v_2+v_3=v_1+v_4$ therefore isolates the failure of velocity composition from the score composition, which is exact at this point.
\end{proof}

\begin{remark}[The counterexample and \cref{thm:and-bias}]
\label{rem:counterexample-interaction}
The four clusters are exactly the four cells of the atom pair, $B_1=c_1\land c_2$, $B_2=c_1\land\lnot c_2$, $B_3=\lnot c_1\land c_2$ and $B_4=\lnot c_1\land\lnot c_2$, with cell-conditional drifts $d_t(a_1,a_2)=2(v_j-5)$ for the corresponding cluster $j$. The mixed second difference \cref{eq:delta-def} is therefore
\begin{align}
        \Delta_t(5)=2\bigl(v_1-v_2-v_3+v_4\bigr)=2\,(0-1-2+10)=14,
\end{align}
and since $\widehat\pi(c_1)=\widehat\pi(c_2)=\tfrac12$, \cref{eq:and-bias-cell} predicts the residual $-\bigl(1-\tfrac12\bigr)\bigl(1-\tfrac12\bigr)\cdot14=-3.5$, matching the direct computation above. Equivalently, by \cref{eq:app-residual-interaction}, $R_t(5)=\tfrac12\cdot\tfrac12\cdot\tfrac12\cdot\tfrac12\cdot14=0.875$ and $-R_t(5)/\bigl(\widehat\pi(c_1)\widehat\pi(c_2)\bigr)=-3.5$. The ``parallelogram'' condition $v_2+v_3=v_1+v_4$ is thus precisely the vanishing of the drift interaction $\Delta_t$, which by \cref{thm:and-bias} is the exact criterion for velocity composition at a conjunction, and which is logically independent of the conditional independence of the atoms holding here.
\end{remark}

\subsection{Consistency of the equilibration}
\label{rem:corrector-invariance}

For fixed $t_{n+1}\in(0,1)$, the continuous-time Langevin dynamics corresponding to \cref{eq:corrector}, driven by the exact composed score $s_{t_{n+1}}(\cdot\mid\varphi)=\nabla_x\log p_{t_{n+1}}(\cdot\mid\varphi)$, admit $p_{t_{n+1}}(\cdot\mid\varphi)$ as an invariant distribution. Under standard regularity and ergodicity assumptions, these dynamics therefore re-equilibrate the sample toward the desired logical marginal, independently of the preceding transport along $\tilde v$. The updates $x_{n+1}^{(k)}$ in \cref{eq:corrector} form the corresponding Euler-Maruyama approximation. Their accuracy improves as $\Delta t_n$ decreases and the effective equilibration time $K\,\bar\gamma(t_{n+1})\,\Delta t_n$ increases. 
In the joint limit $\Delta t_n\to0$ and $K\,\bar\gamma(t_{n+1})\,\Delta t_n\to\infty$, the corrector recovers $p_{t_{n+1}}(\cdot\mid\varphi)$ at that fixed time. At finite $K$, the choice of $\tilde v$ remains important because a well-aligned transport step provides a better initialization for re-equilibration. Taking $\tilde v\equiv 0$ and $K\to\infty$ yields score-only Langevin dynamics at fixed $t$.
In principle, writing \cref{eq:corrector} as a Langevin step with step size $\varepsilon_t=\tfrac12\bar\gamma(t)\,\Delta t$ shows that this coefficient is a free hyperparameter of our method, and could be chosen differently. Unless stated otherwise we use this schedule-tied default. Reasons for this choice and an ablation are given in Appendix \ref{par:langevin_step_size}.

\subsection{Algorithms}
\begin{algorithm}[!h]
\small
\caption{\method: logical sampling with transport--re-equilibration guidance}
\label{alg:logsi}
\begin{algorithmic}[1]
\Require Boolean formula $\varphi$; initial state $x\sim\rho_0$; grids $0=t_0<\cdots<t_N=1$; equilibration budget $K$; flags \textsc{ode}, \textsc{proxy}
\For{$n=0,\ldots,N-1$}
        \State $t\gets t_n$;\quad $\Delta t\gets t_{n+1}-t_n$
        \If{\textsc{proxy}}
                \State $\alpha_i\gets$\textsc{ComposeWeights}$(\varphi, x,t)$ \Comment{\cref{alg:compose-weights}}
                \State $\tilde v_t(x\mid\varphi)\gets v_t(x)+\sum_i \alpha_i^{\varphi}(t,x)\bigl(v_t(x\mid c_i)-v_t(x)\bigr)$ \Comment{\cref{eq:partition-vel}}
        \Else
                \State $\tilde v_t(x\mid\varphi)\gets v_t(x)$ 
        \EndIf
        \If{\textsc{ode}}
                \State $x\gets x+\tilde v_t(x\mid\varphi)\,\Delta t$ \label{line:pred-ode} \Comment{ODE transport}
        \Else
        \State $s_t(x\mid\varphi)\gets$ \textsc{ComposeScore}$(\varphi,x,t)$ \Comment{\cref{alg:compose-score}}
                \State $x\gets x+\bigl(\tilde v_t(x\mid\varphi)+\tfrac{1}{2}\bar\gamma(t)\,s_t(x\mid \varphi)\bigr)\Delta t+\sqrt{\bar\gamma(t)\Delta t}\,\xi$, $\,\xi\sim\mathcal N(0,I)$ \label{line:pred-sde} \Comment{SDE transport}
        \EndIf
        \If{$t_{n+1} \in (0,1)$}   
            \State $t \gets t_{n+1}$
            \For{$k=1,\ldots,K$}
                    \State $s_t(x\mid\varphi)\gets$ \textsc{ComposeScore}$(\varphi,x,t)$ \Comment{\cref{alg:compose-score}}
                    \State $x\gets x+\tfrac{1}{2}\bar\gamma(t)\,s_t(x\mid\varphi)\,\Delta t+\sqrt{\bar\gamma(t)\Delta t}\,\eta$, $\,\eta\sim\mathcal N(0,I)$ \Comment{corrector, \cref{eq:corrector}}
            \EndFor
        \EndIf
        
\EndFor
\State \Return $x$
\end{algorithmic}
\end{algorithm}

\begin{algorithm}[!h]
\caption{\textsc{ComposeWeights}: recursive LogDiff coefficients $(\alpha_i^{\varphi})_{i=1}^{k}$ (\cref{sec:prelim-logdiff})}
\label{alg:compose-weights}
\begin{algorithmic}[1]
\Require Boolean formula $\varphi$ over atoms $\{c_i\}_{i=1}^{k}$; state $x$; time $t$; classifier posteriors $\widehat\pi(\cdot)=p_t(\cdot\mid x)$
\Ensure coefficients $\{\alpha_i^{\varphi}(t,x)\}_i$ such that $s_t(\varphi\mid x)=\sum_i\alpha_i^{\varphi}\,s_t(c_i\mid x)$
\State $(\alpha^{\varphi},\,\widehat\pi(\varphi))\gets\textsc{LogicalWeights}(\varphi)$
\State \Return $\alpha^{\varphi}$
\Statex
\Function{LogicalWeights}{$\psi$} \Comment{returns $(\alpha^{\psi},\widehat\pi(\psi))$ with $\alpha^{\psi}\in\mathbb R^{k}$}
        \If{$\psi=c_j$ is atomic}
                \State $\alpha\gets e_j$ \Comment{one-hot: $\alpha_j=1$, $\alpha_{i\neq j}=0$}
                \State \Return $(\alpha,\;\widehat\pi(c_j))$
        \ElsIf{$\psi=\lnot\chi$}
                \State $(\alpha_\chi,\pi_\chi)\gets\textsc{LogicalWeights}(\chi)$
                \State \Return $\bigl(-\tfrac{\pi_\chi}{1-\pi_\chi}\,\alpha_\chi,\;1-\pi_\chi\bigr)$ \Comment{NOT}
        \ElsIf{$\psi=\chi\land\xi$}
                \State $(\alpha_\chi,\pi_\chi)\gets\textsc{LogicalWeights}(\chi)$;\; $(\alpha_\xi,\pi_\xi)\gets\textsc{LogicalWeights}(\xi)$
                \State \Return $(\alpha_\chi+\alpha_\xi,\;\pi_\chi\pi_\xi)$ \Comment{AND}
        \ElsIf{$\psi=\chi\lor\xi$ and mutually exclusive}
                \State $(\alpha_\chi,\pi_\chi)\gets\textsc{LogicalWeights}(\chi)$;\; $(\alpha_\xi,\pi_\xi)\gets\textsc{LogicalWeights}(\xi)$
                \State $\pi\gets\pi_\chi+\pi_\xi$
                \State \Return $\bigl(\tfrac{\pi_\chi}{\pi}\alpha_\chi+\tfrac{\pi_\xi}{\pi}\alpha_\xi,\;\pi\bigr)$ \Comment{OR-ME}
        \ElsIf{$\psi=\chi\lor\xi$ and conditionally independent}
                \State $(\alpha_\chi,\pi_\chi)\gets\textsc{LogicalWeights}(\chi)$;\; $(\alpha_\xi,\pi_\xi)\gets\textsc{LogicalWeights}(\xi)$
                \State $\pi\gets\pi_\chi+\pi_\xi-\pi_\chi\pi_\xi$
                \State $\alpha\gets\tfrac{\pi_\chi(1-\pi_\xi)}{\pi}\alpha_\chi+\tfrac{\pi_\xi(1-\pi_\chi)}{\pi}\alpha_\xi$
                \State \Return $(\alpha,\;\pi)$ \Comment{OR-CI}
        \EndIf
\EndFunction
\end{algorithmic}
\end{algorithm}

\begin{algorithm}[!h]
\caption{\textsc{ComposeScore}: LogDiff score composition via composed weights (\cref{sec:prelim-logdiff,app:recursive_scores})}
\label{alg:compose-score}
\begin{algorithmic}[1]
\Require Boolean formula $\varphi$; state $x$; time $t$; atomic networks $s_t(x)$, $s_t(x\mid c_i)$
\Ensure composed conditional score $s_t(x\mid\varphi)=\nabla_x\log p_t(x\mid\varphi)$
\State $\{\alpha_i^{\varphi}(t,x)\}_i\gets\textsc{ComposeWeights}(\varphi,x,t)$ \Comment{\cref{alg:compose-weights}}
\State \Return $s_t(x)+\sum_{i=1}^{k}\alpha_i^{\varphi}(t,x)\bigl[s_t(x\mid c_i)-s_t(x)\bigr]$ \Comment{\cref{eq:logdiff_lin_comb_score_xphi}}
\end{algorithmic}
\end{algorithm}
\clearpage

\subsection{Relation to SuperDiff}
\label{app:skreta}
SuperDiff~\cite{skreta:2025} introduce a method to track changes in the conditional densities $p_t(x\mid c_i)$ along a sample path without training a noise-aware classifier. Their Theorem~1 gives an It\^o estimator for $\mathrm d\log p_t(x\mid c_i)$.
In principle, this construction could also be applied to stochastic interpolants and bridges as in our setting, although they only develop it for diffusion models.

For logical \texttt{AND}, they choose weights $\kappa_i$ (with $\sum_i\kappa_i=1$) by requiring equal infinitesimal density updates, $\mathrm d\log p_t(x\mid c_i)=\mathrm d\log p_t(x\mid c_j)$, and transport the state with a $\kappa$-weighted linear combination of the atomic SDE drifts, so that in expectation the endpoint is equally likely under each conditional model, $p_1(x\mid c_i)=p_1(x\mid c_j)$ for all $i,j$.
Crucially, matching the \emph{increments} $\mathrm d\log p_t(x\mid c_i)$ along the trajectory only pins down terminal densities \emph{up to the initial gap}: integrating from $0$ to $1$ gives
\begin{align}
\label{eq:superdiff-endpoint-gap}
\log p_1(x_1\mid c_i)-\log p_1(x_1\mid c_j) \;=\; \log p_0(x_0\mid c_i)-\log p_0(x_0\mid c_j),
\end{align}
so the construction lands on the \texttt{AND} manifold $\{x:p_1(x\mid c_i)=p_1(x\mid c_j)\ \forall i,j\}$ only when the initial laws already agree, $p_0(x\mid c_i)=p_0(x\mid c_j)$ for all $i,j$.
This is true for standard diffusion models, where generation begins from a shared Gaussian independent of the condition, but fails in our bridge setting: we do not start from noise but from explicit samples $x_0\sim\rho_0$, and in general $p_0(x\mid c_i)\neq p_0(x\mid c_j)$. By \eqref{eq:superdiff-endpoint-gap}, the endpoint gap then equals the initial gap and the resulting sample is not on the \texttt{AND} manifold. 

A further practical difference concerns the divergence term in their density update.
For VP/OU diffusion, $\langle\nabla,f_t\rangle$ in Theorem~1 is available in closed form because the forward drift $f_t$ is linear.
For general stochastic interpolants one would instead need $\operatorname{div}$ of each atomic drift $v_t(\cdot\mid c_i)+\tfrac12\bar\gamma(t)\,s_t(\cdot\mid c_i)$, which for neural $v_t,s_t$ must be estimated stochastically at every step (e.g.\ via Hutchinson's estimator), adding per-step cost and variance to the density updates rather than being a fundamental obstruction.

For these reasons we do not compare against SuperDiff in our experiments. Adapting their density-tracking construction to bridge interpolants, either by re-deriving the \texttt{AND} weights so that \eqref{eq:superdiff-endpoint-gap} is closed by the choice of $\kappa$, or by initialising from a coupling under which the initial gap vanishes, is a natural future direction.

\section{Experimental Details}
\label{sec:exp-details}

\subsection{Training hyperparameters}
\label{sec:exp-hparams}

Training hyperparameters for the velocity and score networks are summarised in \cref{tab:train-hparams}, architecture and evaluation-metric details follow in subsequent subsections.

\begin{table}[!h]
\caption{\textbf{Training hyperparameters} for our experiments. CFG dropout means the probability with which the condition is exchanged with the unconditional symbol.}
\label{tab:train-hparams}
\begin{center}
\begin{tabular}{lccc}
\toprule
\multicolumn{1}{c}{\bf Hyperparameter} & \multicolumn{1}{c}{\bf CMNIST} & \multicolumn{1}{c}{\bf CelebA} & \multicolumn{1}{c}{\bf QM9} \\
\midrule
Epochs & 500 & 500 & 500 \\
Batch size & 1024 & 64 & 128 \\
Learning rate & $10^{-3}$ & $2\times10^{-4}$ & $10^{-4}$ \\
Weight decay & $10^{-5}$ & $0.01$ & $0.01$ \\
Noise scheduler $\gamma(t)$ & $4\,t(1-t)$ & $8\,t(1-t)$ & $t(1-t)$\\
Optimizer / scheduler & \multicolumn{3}{c}{AdamW + cosine annealing to $0$ over training} \\
CFG dropout & \multicolumn{3}{c}{$20\%$ per-label / per-atom unconditional tokens} \\
\bottomrule
\end{tabular}
\end{center}
\end{table}

\subsection{Datasets and transfer protocols}
\label{app:datasets}
\paragraph{EMNIST to CMNIST.}
The source distribution $\rho_0$ is the EMNIST dataset restricted to ten letter classes (\texttt{a-e}, \texttt{A-E}), the target distribution $\rho_1$ is the Colored MNIST dataset with ten digit and ten colour classes (10 colours are multiplied by the greyscale MNIST samples). Images are rendered as $28{\times}28$ RGB tensors in $[0,1]$. Each sample is initialised from a concrete source image and transported toward CMNIST while enforcing a Boolean formula $\varphi$ over \texttt{Digit}$(d)$ and \texttt{Color}$(c)$ atoms. We evaluate fixed templates (AND, OR\textsubscript{ME}, OR\textsubscript{CI}, AND$\neg$) and randomly generated expression trees of depth two--three over the same atom vocabulary, while ensuring that conditional independence and mutual exclusivity hold. Cross-axis formulas such as \texttt{Digit}$(d)\lor\texttt{Color}(c)$ use the OR-CI rule of \cref{app:recursive_scores}, while same-axis pairs such as \texttt{Digit}$(d_1)\lor\texttt{Digit}(d_2)$ (or the analogue for colours) use OR-ME.

\paragraph{CelebA.}
We train a stochastic interpolant from CelebA \emph{female} to \emph{male} images in the latent space of the FLUX.1-schnell VAE ($16$ channels, $16{\times}16$ spatial resolution). Logical atoms are two binary CelebA attributes: \texttt{Glasses} and \texttt{Hat}. We test the logical statements AND, OR-CI, and AND$\neg$. Decoded outputs are evaluated at $128{\times}128$ in pixel space.

\paragraph{QM9.}
We bridge noisy \emph{source} molecules drawn from the QM9 dataset (matched in atom count to each target) to target molecules from QM9. Each graph carries $3$D coordinates and a $5$-way atom-type vector over $\{\mathrm{H},\mathrm{C},\mathrm{N},\mathrm{O},\mathrm{F}\}$, coordinates are center-of-mass centred. Atoms are binary molecular attributes (ring presence, oxygen presence, etc.) composed with the same LogDiff rules as in \cref{sec:prelim-logdiff}. The main logical statement pairs \texttt{has\_rings}$=1$ with \texttt{has\_oxygen}$=1$ under AND, OR-CI, and AND$\neg$. Bridge sources are restricted to molecules whose ground-truth labels \emph{violate} the rule, so conformity measures whether the sampler can \emph{create} the constraint from a non-conforming start.

\subsection{Model architectures}
\label{sec:exp-architectures}

\begin{table}[t]
\caption{\textbf{Classifier accuracies} for the metrics used in the CMNIST and CelebA experiments.}
\begin{center}
\begin{tabular}{llc}
\toprule
\textbf{Dataset} & \textbf{Attribute} & \textbf{Accuracy} \\
\midrule
\multirow{3}{*}{CMNIST}
  & Digit & 0.996 \\
  & Color & 1.000 \\
  & Joint & 0.996 \\
\midrule
\multirow{3}{*}{CelebA}
  & Glasses & 0.993 \\
  & Hat & 0.986 \\
  & Joint & 0.979 \\
\bottomrule
\end{tabular}
\end{center}
\label{tab:classifier-acc}
\end{table}

Each benchmark uses a separate trained velocity network $v_\theta$, and score network $s_\theta$. A time-aware classifier for posterior weights in OR / negation, and a held-out evaluator for the conformity metric is also trained (the molecular experiments uses rdkit as conformity metric). All the models were trained in a CFG way, to directly predict unconditioned and conditioned output with the same model.

\subsubsection{EMNIST to CMNIST}

\paragraph{Velocity and score fields.}
Both fields are U-Net-style ResNets with base width $96$, conditioning dimension $64$, and $3$-channel $28{\times}28$ inputs. The encoder-decoder has three downsample stages (average pooling) and skip connections; residual blocks use FiLM-style modulation\citep{perez2018film} from the combined $(\text{digit},\text{colour},t)$ embedding. The velocity head outputs a $3$-channel field, the score head outputs a $3$-channel score and, on its bottleneck features, two auxiliary classifiers (digit and colour logits) used at sampling time.

\paragraph{Noise-aware classifier (guidance).}
During sampling, atom log-probabilities $\widehat\pi(c_i\mid x_t)$ for disjunction and negation are read from the \emph{score} network's auxiliary heads: the model is queried with the unconditional wildcard label on both axes, and log-softmax probabilities are taken for the atom's digit or colour value at the current $(x_t,t)$. No separate guidance classifier is trained for CMNIST.

\paragraph{Conformity evaluator.}
Rule conformity on generated images is measured with a held-out classifier: ResNet18 adapted for $28{\times}28$ inputs ( $3{\times}3$ stem, no initial max-pool) with separate linear heads for digit and colour. This model is trained only on clean Colored MNIST and is \emph{not} used inside the sampler. Accuracy of the classifier is reported in \cref{tab:classifier-acc}.
\subsubsection{CelebA}

\paragraph{Velocity and score fields.}
Both fields are class-conditional SiT transformers trained on $16{\times}16$ VAE latents with $16$ input channels, patch size $1$, hidden size $384$, depth $12$, and $6$ attention heads. CelebA's $40$ binary attributes are embedded with a multi-label encoder (per-attribute $3$-way tables: negative / positive / wildcard) and summed with a sinusoidal timestep embedding. Outputs are unpatchified to a $16$-channel velocity or score field in latent space.

\paragraph{Noise-aware classifier (guidance).}
OR and AND$\neg$ require time-conditioned atom posteriors on the \emph{noisy latent} $x_t$. We train a SiT model on the same latent geometry (patch size $2$, hidden size $192$, depth $6$, $6$ heads) with timestep conditioning only. The model outputs $40$ independent logits; at step $t$ the probability that attribute $a$ equals value $v$ is $\sigma(\ell_a)$ if $v{=}1$ and $1-\sigma(\ell_a)$ if $v{=}0$. With probability $0.1$ during classifier training, $t{=}0$ and clean latents are shown so the network is calibrated near the source endpoint.

\paragraph{Conformity evaluator.}
After VAE decoding to $128{\times}128$ RGB, conformity uses a ResNet classifier trained on male CelebA samples with a $40$-way sigmoid head (as a training starting point the ResNet classifier was initialized with the ImageNet trained weights). Accuracy of the classifier is reported in \cref{tab:classifier-acc} (Only the tested attributes are reported).

\subsubsection{QM9}

\paragraph{Velocity and score fields.}
Both fields are EGNNs built on fully connected molecular graphs (all ordered atom pairs within each molecule). We use the EDM-stabilised EGNN variant with hidden size $256$, depth $9$, and $5$ atom-type channels. Node inputs are continuous one-hot atom vectors and $3$D coordinates; a conditioning module adds a sinusoidal time embedding plus embeddings for the binary graph labels (\texttt{has\_rings}$=1$, \texttt{has\_oxygen}$=1$ etc.), each with an extra ``unknown'' index for classifier-free guidance. The network predicts equivariant position velocities and invariant atom-type velocities; discrete types are recovered by argmax at the end of sampling.

\paragraph{Noise-aware classifier (guidance).}
The classifier shares the same graph construction but is a classification-only stack (hidden size $256$, depth $6$): equivariant message-passing blocks, mean pooling over nodes, layer norm, and a linear head producing one logit per binary attribute. It is trained on noisy graphs $(x_t,h_t)$ with the same bridge noise as the generator; atom posteriors for OR / AND$\neg$ are sigmoids of the corresponding logits at the current $t$.

\paragraph{Validity and conformity.}
Chemical validity and rule satisfaction are evaluated with RDKit (not with a learned classifier like in the image cases). Conformity is defined only on RDKit-valid molecules.

\subsection{Evaluation metrics}
\label{sec:exp-metrics}

\subsubsection{Image benchmarks}

\paragraph{Rule conformity.}
For a batch with logical formula $\varphi$, conformity is the fraction of generated images whose \emph{evaluator} predictions satisfy $\varphi$. On CMNIST and CelebA the evaluator is the ResNet18 classifier, see last section. This is distinct from the noise-aware classifiers used inside the sampler.

\paragraph{FID.}
We report Fr\'echet Inception Distance (FID) with Inception-v3 \texttt{pool3} features ($2048$-dimensional, \texttt{torchmetrics} implementation, \texttt{normalize=True}). Generated and reference images are bilinearly resized to $128{\times}128$ and clamped to $[0,1]$ before feature extraction.

For CMNIST, FID is computed \emph{per expression}: for each $\varphi$ in the evaluation schedule, the reference set is the subset of real Colored MNIST training images whose ground-truth digit and colour satisfy $\varphi$; generated samples for that batch are compared to this expression-conditioned reference (up to $\min(10{,}000,\lvert\text{subset}\rvert)$ real images). The reported FID is the mean across expression batches.

For CelebA, it is similar. We report the FID scores with the expression-conditioned references (up to $5{,}000$ reals).

\subsubsection{QM9 metrics}
\label{sec:exp-qm9-metrics}

\paragraph{RDKit validity.}
For each molecule we build an RDKit conformer from predicted coordinates and atomic numbers, infer bonds, and run \texttt{SanitizeMol}. The \emph{validity} rate is the fraction of samples for which this pipeline succeeds.

\paragraph{Rule conformity.}
On valid molecules only, we recompute binary attributes from the perceived structure: ring-based labels via RDKit ring perception (\texttt{has\_rings}, etc.) and element-presence labels by atomic number. Conformity is the fraction of valid molecules whose recomputed attributes satisfy $\varphi$ (AND / OR / AND$\neg$ over the two chosen atoms).

\paragraph{Joint success.}
The primary success metric reported in the main figures is $\text{validity}\times\text{conformity}$ (equivalently, the fraction of \emph{all} samples that are both valid and rule-conforming).

\paragraph{Structural edit distance (RMSD).}
To calculate the distance between source and final molecule, we first apply the Kabsch algorithm \citep{kabsch1976solution} after centering both structures at zero centre of mass, removing rigid rotation ambiguity, and second calculate the RMSD between the positions of the atoms. As a baseline, we calculate the mean Kabsch-RMSD between two random QM9 molecules. This is purely a geometric metric, the atomtypes are not taken into account.

\paragraph{Novelty.}
Valid molecules are converted to canonical SMILES. \emph{Novelty} is the fraction of \emph{unique} valid SMILES not present in a reference set of canonical SMILES for the full QM9 dataset. 

\subsubsection{Compute matching and error bars}
\label{sec:exp-compute-matching}

Network-evaluation (NFE) budgets for \method\ vs.\ matched baselines follow \cref{sec:experiments}: outer steps $N$ with $K$ corrector substeps are matched against a $K{=}0$ baseline with additional integration steps, such that the number of evaluations of the velocity and score networks match. When multiple training or evaluation seeds are run, main-text error bars aggregate across seeds for training and evaluation.

\subsection{Additional Experiments}

\paragraph{Langevin step size ablation}\label{par:langevin_step_size}
The Langevin corrector step size $\varepsilon$ is in principle a free hyperparameter of \method.
In the main experiments we use the schedule-tied choice
\begin{equation}\label{eq:langevin_step_size}
        \varepsilon_t=\tfrac12\bar\gamma(t)\,\Delta t,
\end{equation}
which reuses the SI noise scheduler, stays consistent with the $\Delta t$-based equilibration limit in the main text, and introduces no extra hyperparameter.
As an ablation we instead use the variance-exploding (VE) signal-to-noise (SNR) rule of \citet{song2021scorebased}, originally proposed for Langevin correctors in score-based \emph{diffusion} sampling:
\begin{align}\label{eq:langevin_step_size_snr}
        \varepsilon = 2\Bigl(r\,\frac{\|z\|}{\|s_t(x\mid\varphi)\|}\Bigr)^2,\qquad
        x \leftarrow x+\varepsilon\,s_t(x\mid\varphi)+\sqrt{2\varepsilon}\,z,
        \qquad z\sim\mathcal N(0,I),
\end{align}
with target SNR $r>0$.

\begin{figure}[!h]
    \centering
    \includegraphics[width=0.8\textwidth]{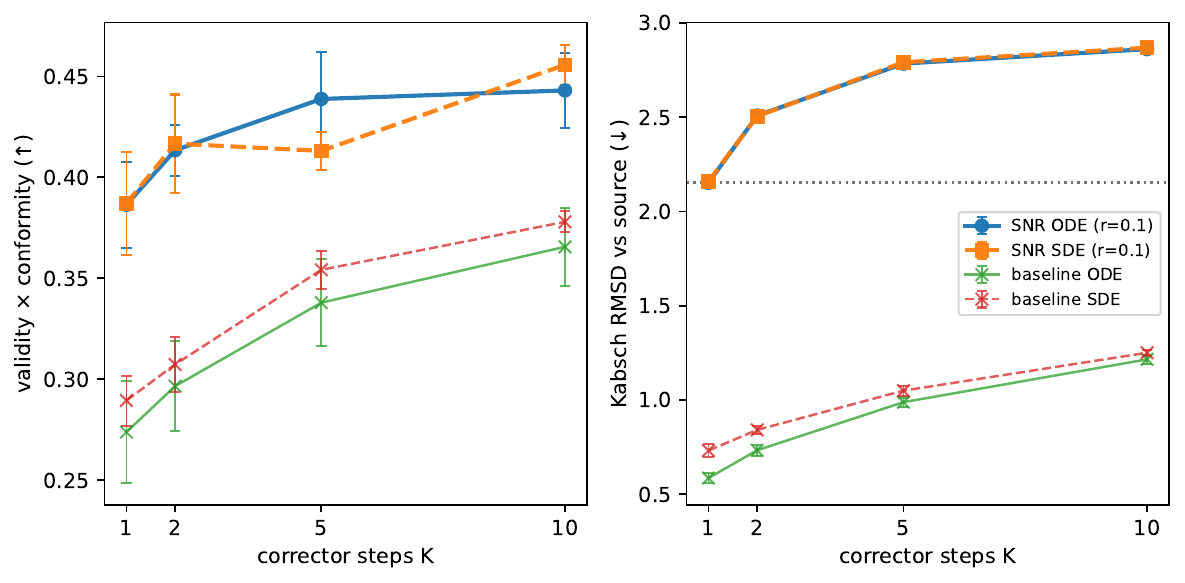}
    \caption{%
        \textbf{Langevin step size ablation on QM9.}
        Left: conformity, right: Kabsch distance to the source as a function of correction steps.
        We compare schedule-tied steps \cref{eq:langevin_step_size} to the SNR rule of \citet{song2021scorebased} \cref{eq:langevin_step_size_snr}, for both ODE and SDE transport.
    }
    \label{fig:additional_langevin_steps}
\end{figure}

We apply the same rule in our bridge / stochastic-interpolant sampler, where the corrector targets the logically composed marginal $p_t(\cdot\mid\varphi)$ rather than a reverse diffusion.
Because this choice introduces the additional hyperparameter $r$, we grid-search $r\in\{0.05,0.1,0.16,0.2\}$ and report the value $r=0.1$ with highest conformity in the case of the QM9 dataset.
\Cref{fig:additional_langevin_steps} shows that SNR tuning can raise logical conformity, but the resulting steps are often large enough that source identity is effectively lost (Kabsch RMSD approaching the level of unrelated source--target pairs).
In the QM9 editing setting this is undesirable, so we keep the schedule-tied step size as the default: it trades a possible conformity gain for edits that remain local to the source. Designing step sizes that improve conformity without erasing source identity is left to future work.

\begin{figure}[!h]
    \centering
    \includegraphics[width=1.0\textwidth]{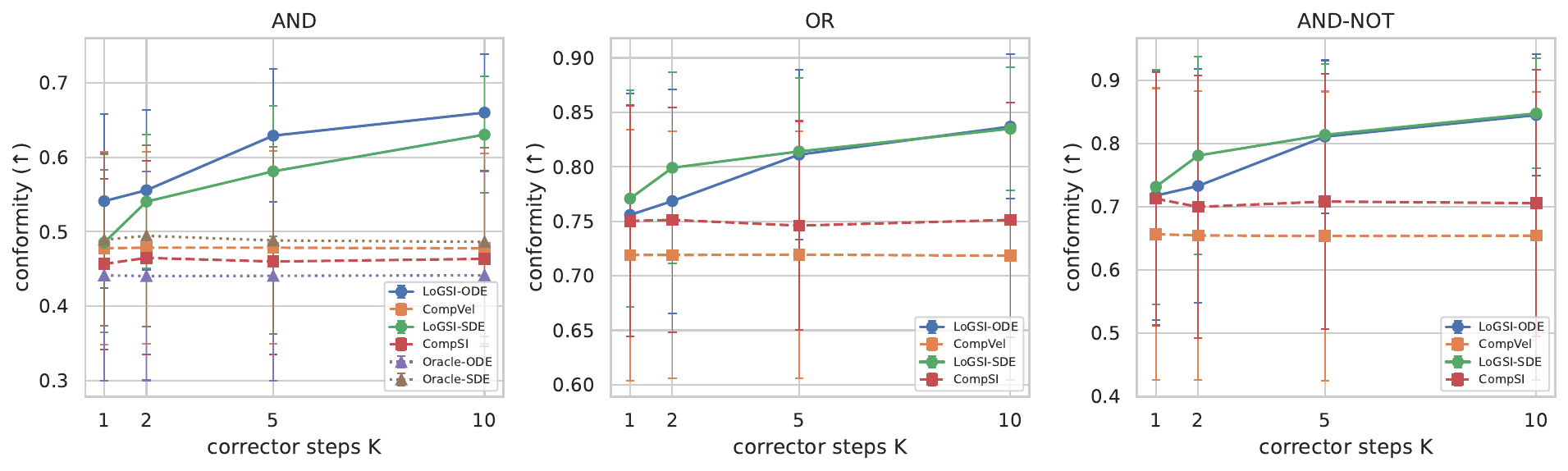}
    \includegraphics[width=1.0\textwidth]{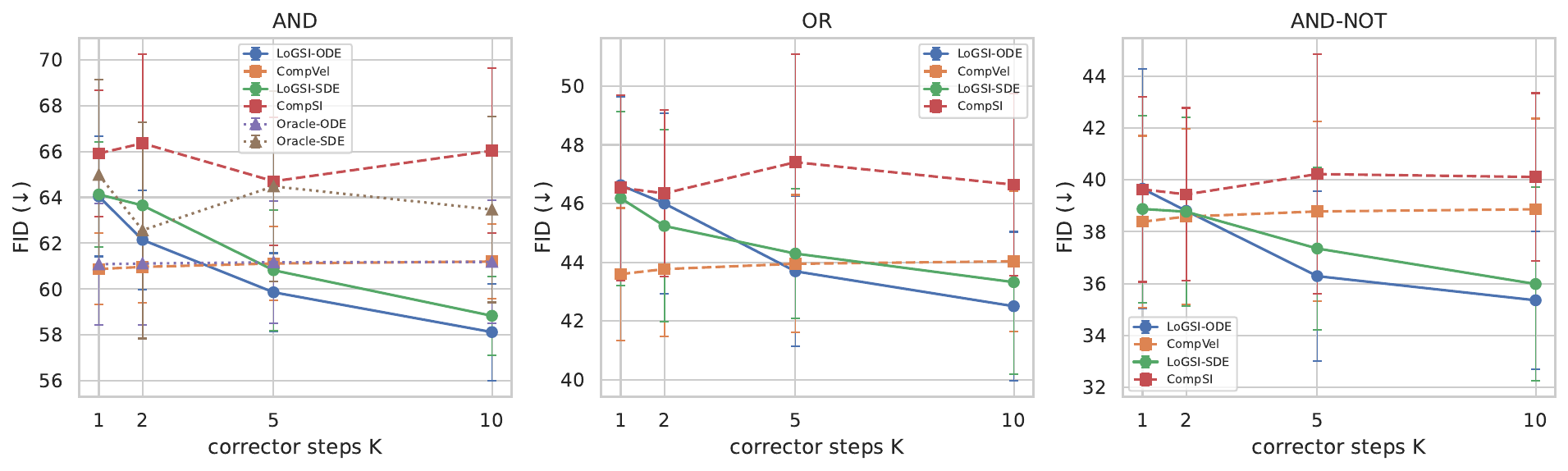}
    \includegraphics[width=1.0\textwidth]{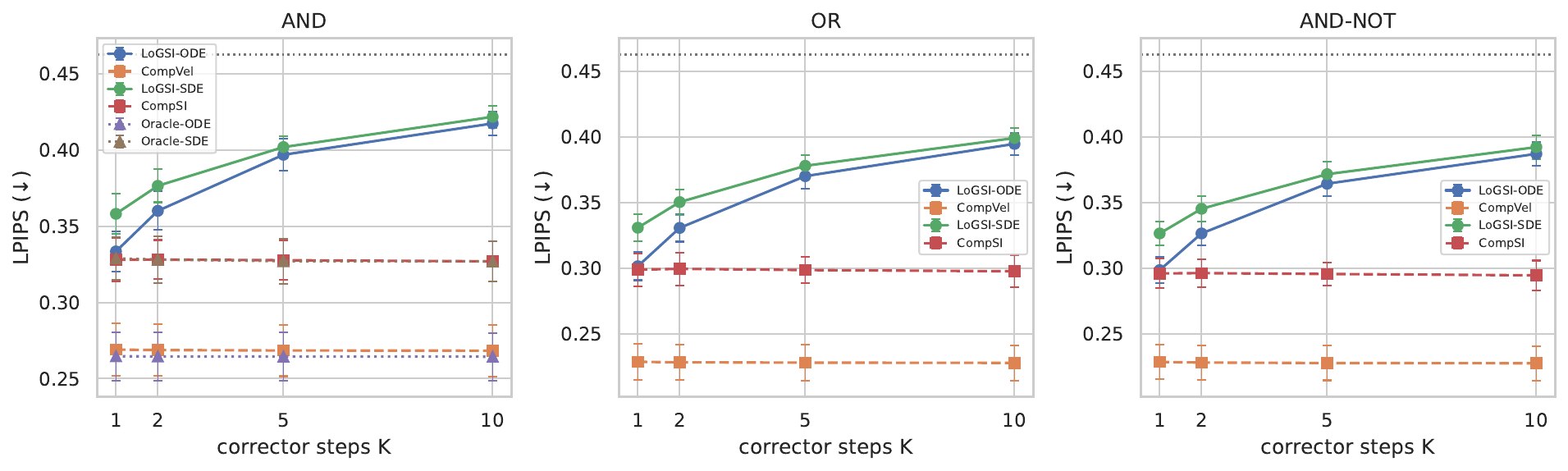}
    \caption{%
        \textbf{Additional CelebA experiments.} Further metrics of the experiment regarding transfer learning on CelebA. We report conformity, FID score and additionally we report LPIPS distance to the starting sample. The dotted horizontal is the mean random LPIPS score between female source and male target sample in the CelebA dataset.
    }
    \label{fig:additional_celeba_results}
\end{figure}

\begin{figure}[!h]
    \centering
    \includegraphics[width=1.0\textwidth]{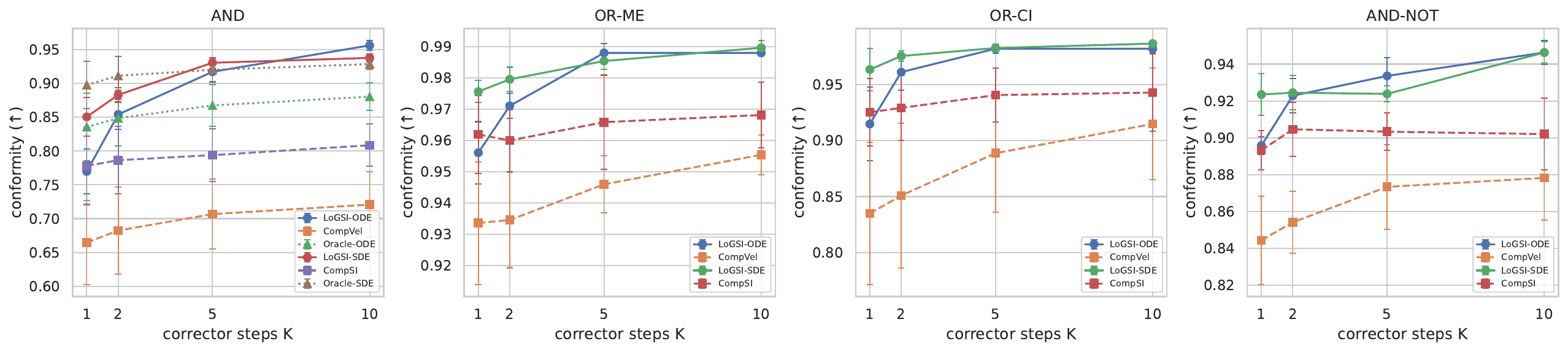}
    \includegraphics[width=1.0\textwidth]{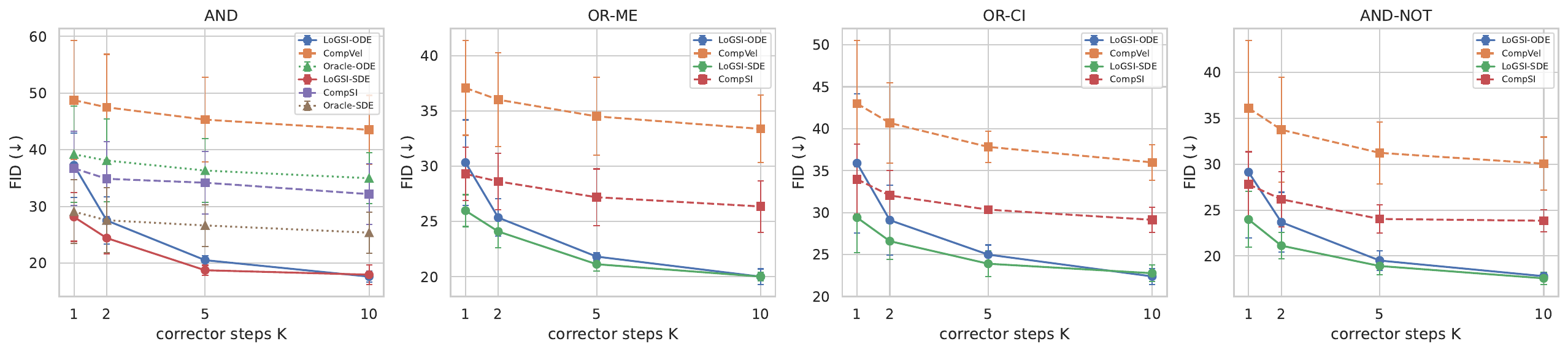}
    \includegraphics[width=1.0\textwidth]{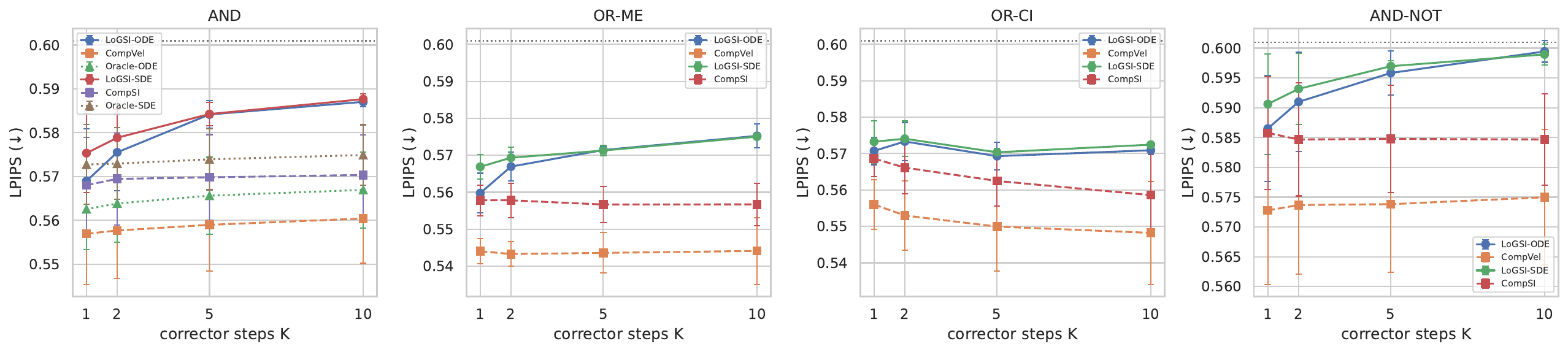}
    \includegraphics[width=0.8\textwidth]{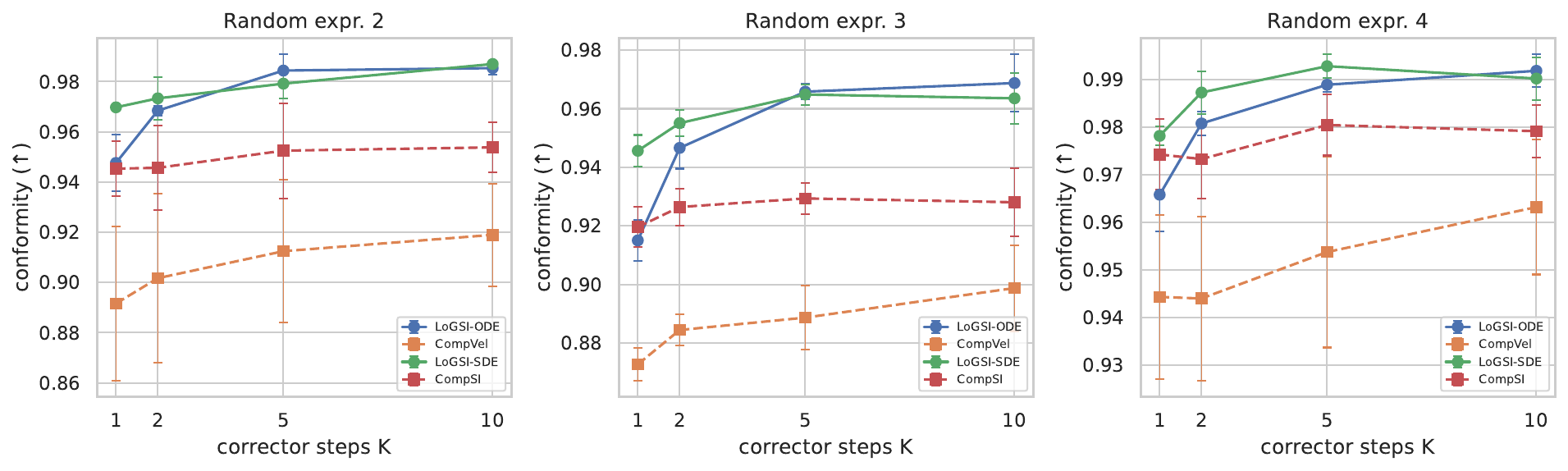}
    \includegraphics[width=0.8\textwidth]{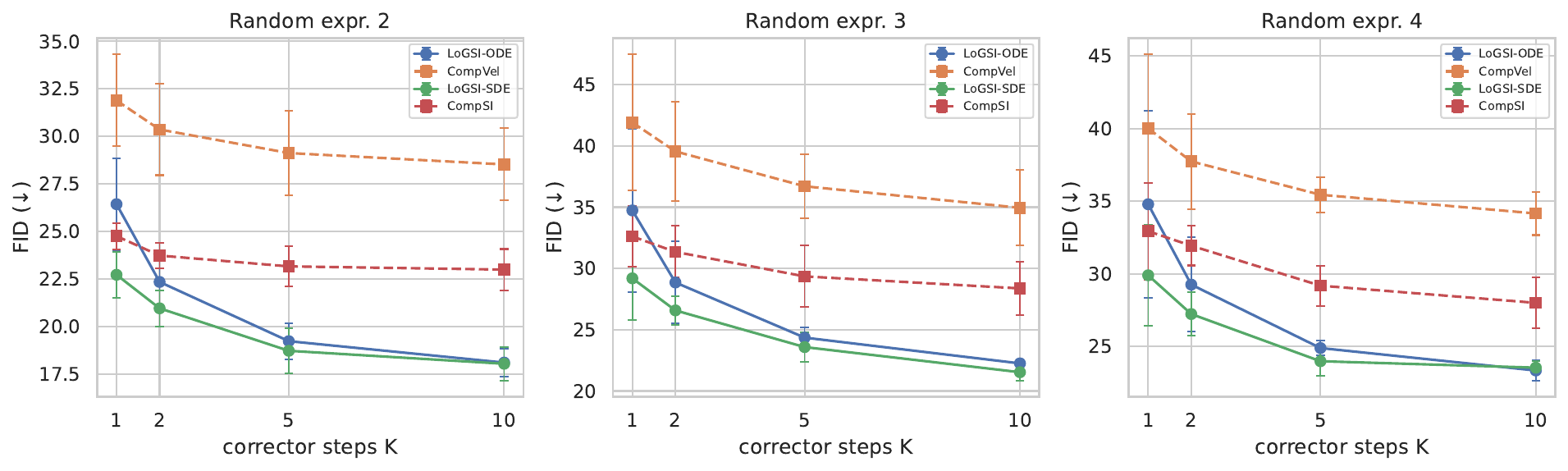}
    \includegraphics[width=0.8\textwidth]{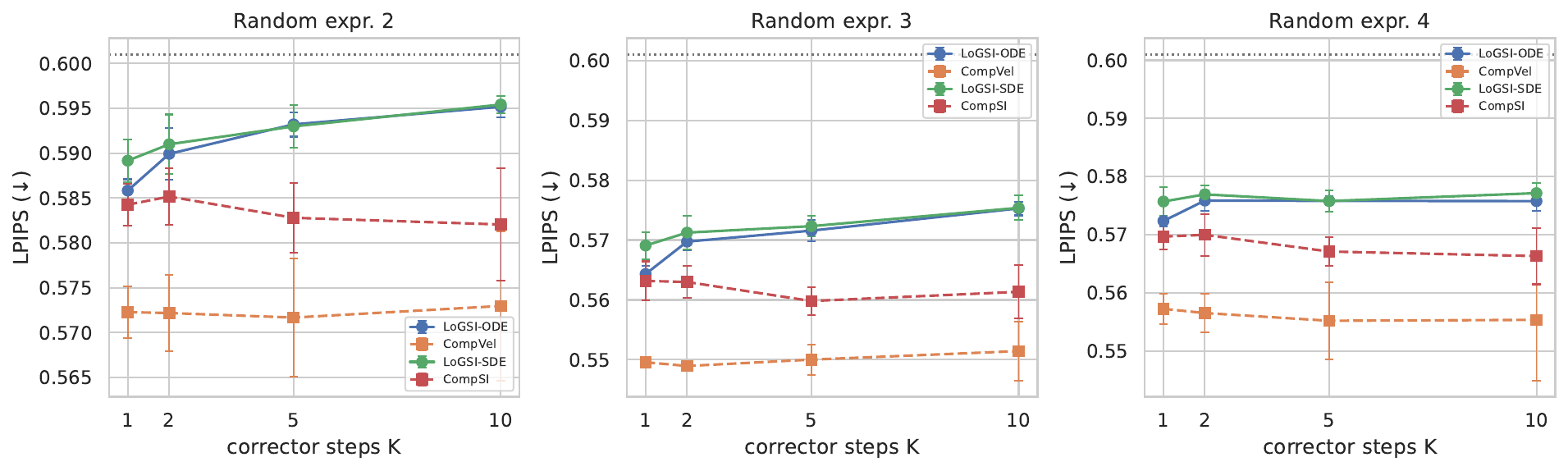}
    \caption{%
        \textbf{Additional CMNIST experiments.} Further metrics of the experiment regarding transfer learning from EMNIST to CMNIST. We report FID score and additionally we report LPIPS distance to the starting sample. The dotted horizontal is the mean random LPIPS score between source sample in EMNIST and target sample in the CMNIST dataset.
    }
    \label{fig:additional_cmnist_results}
\end{figure}

\begin{figure}[!h]
    \centering
    \includegraphics[width=1.0\textwidth]{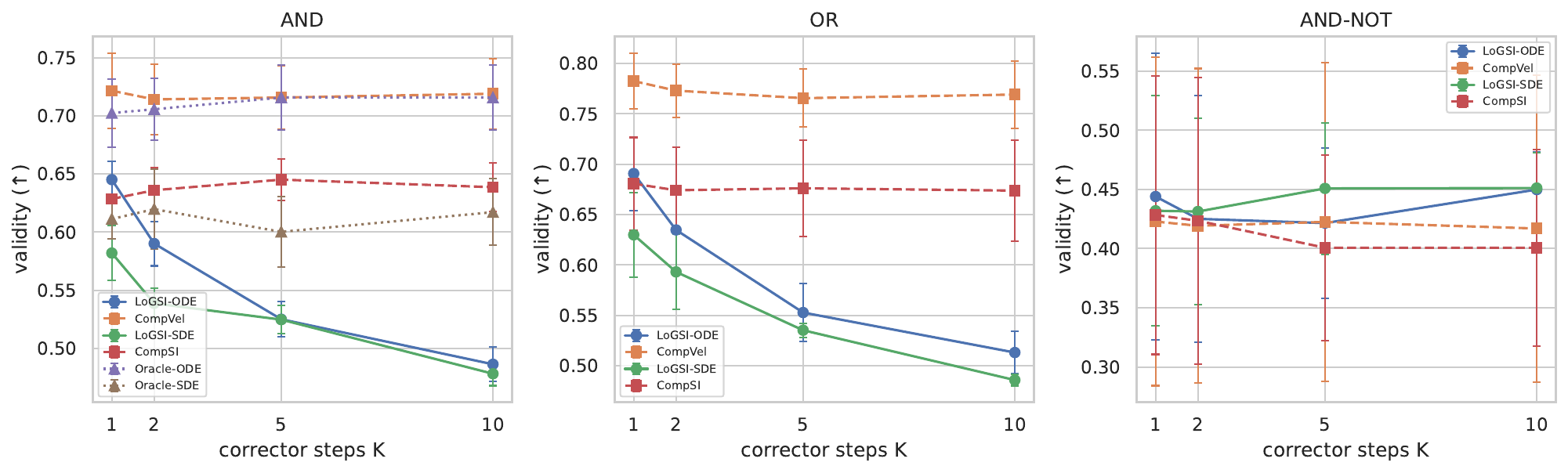}
    \includegraphics[width=1.0\textwidth]{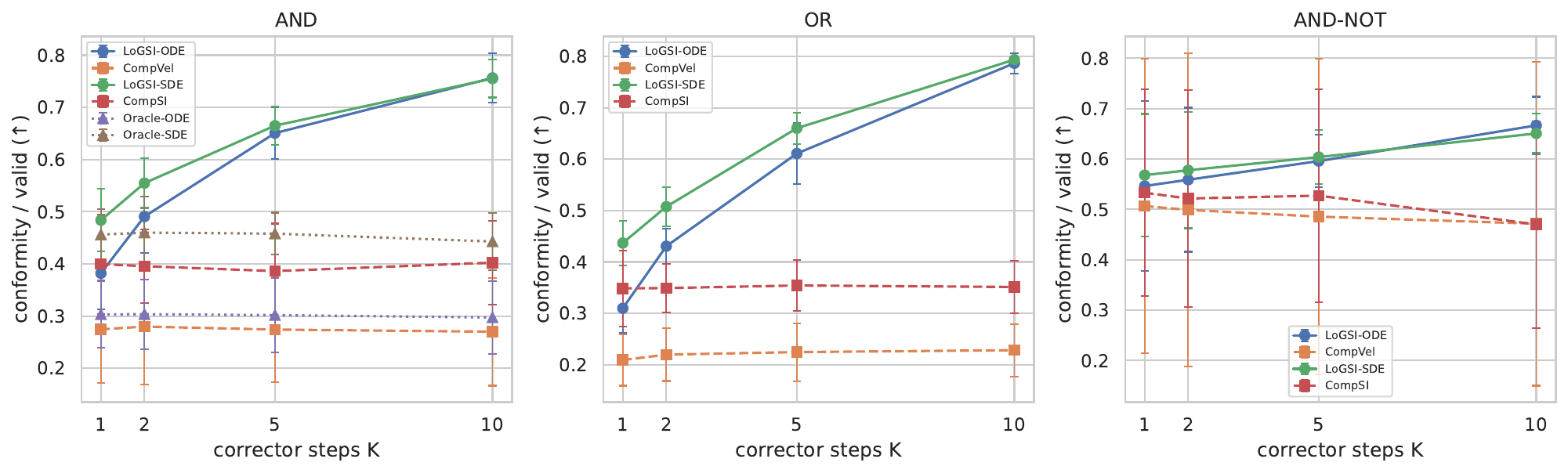}
    \includegraphics[width=1.0\textwidth]{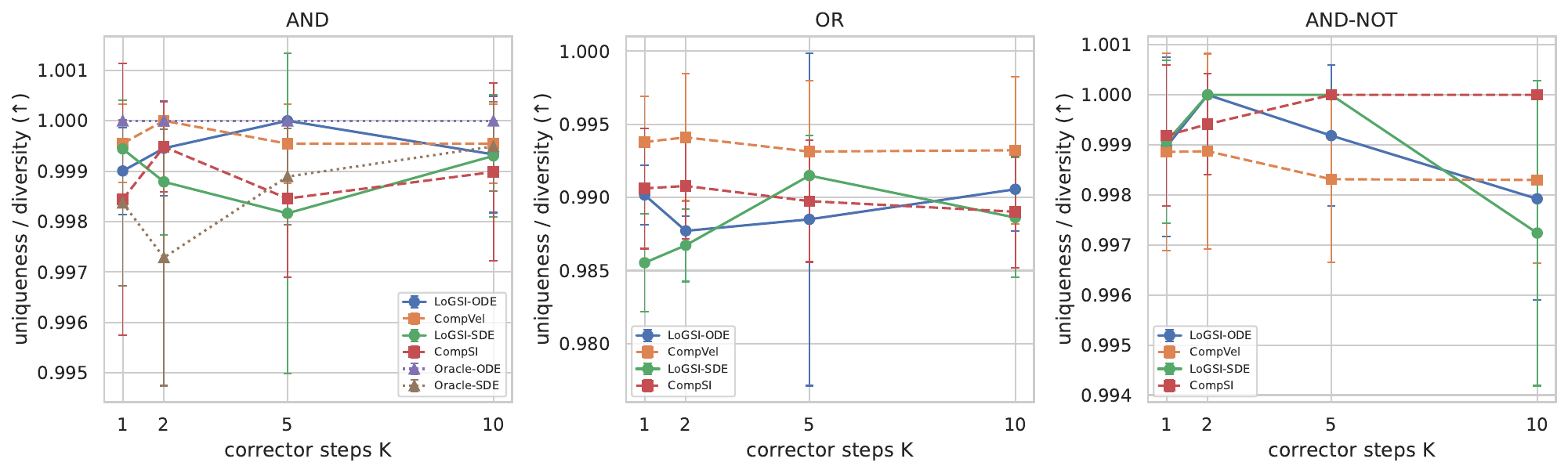}
    \includegraphics[width=1.0\textwidth]{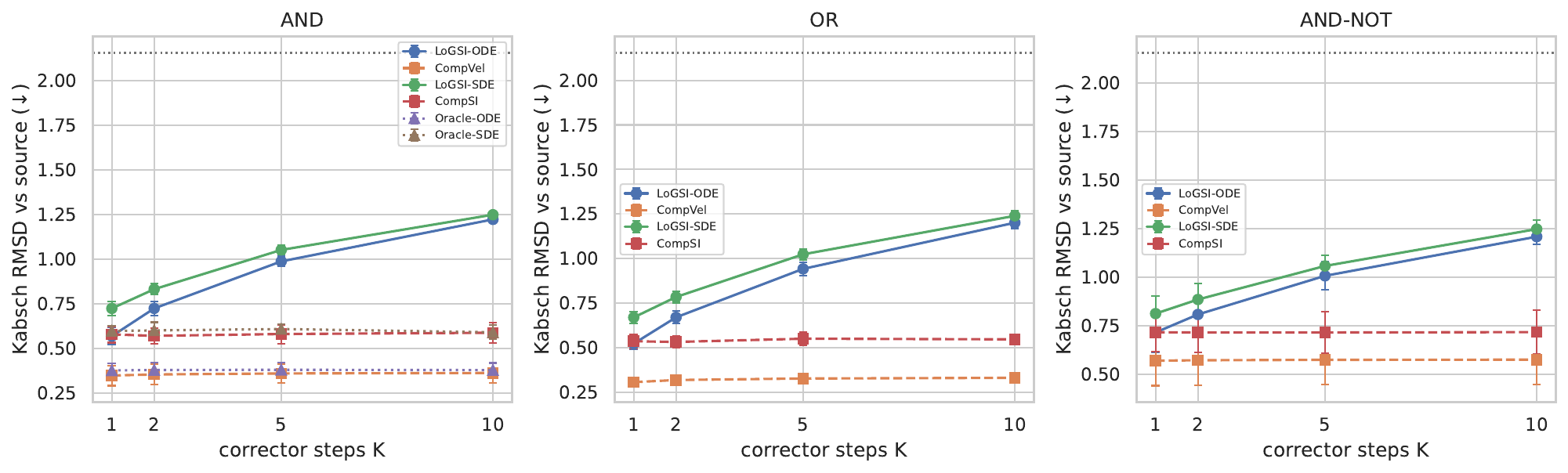}
    \caption{%
        \textbf{Additional QM9 experiments.} Further metrics of the experiment regarding QM9 editing. We report validity and conformity separately. Furthermore, we report uniqueness and Kabsch RMSD. The dotted horizontal is the mean random Kabsch RMSD score between QM9 samples.
    }
    \label{fig:additional_qm9_results}
\end{figure}

\begin{table}[t]
\caption{\textbf{Transport velocity ablation.} 1 step ablation of the experiment given in \cref{tab:transport_ab_ode}}
\label{tab:transport_ab_ode2}
\begin{center}
\begin{tabular}{lccc}
\toprule
\multicolumn{1}{c}{\bf Method}  &\multicolumn{1}{c}{\bf CMNIST}&\multicolumn{1}{c}{\bf CelebA}&\multicolumn{1}{c}{\bf QM9} 
\\ \midrule
\method\ 1 step        & $\mathbf{0.79 \pm 0.03}$ & $\mathbf{0.54 \pm 0.12}$ & $\mathbf{0.27 \pm 0.03}$\\
uncd. 1 step           & $0.020 \pm 0.001$ & $0.016 \pm 0.009$ & $0.15 \pm 0.02$\\
$\tfrac{1}{2}v_t(x|c_1)+\tfrac{1}{2}v_t(x|c_2)$ 1 step            & $0.44 \pm 0.02$ & $0.16 \pm 0.05$ & $0.20 \pm 0.01$\\
only score 1 step      & $0.033 \pm 0.002$ & $0.018 \pm 0.006$ & $0.016 \pm 0.005$\\[2pt]

\bottomrule
\end{tabular}
\end{center}
\end{table}

\begin{table}[!h]
\caption{\textbf{Transport velocity ablation.} Ablations on each of the proposed experiments, where the transport step velocity is exchanged by unconditional velocity, mean velocity between the two conditions and not using a transport step.}
\label{tab:transport_ab_sde}
\begin{center}
\begin{tabular}{lccc}
\toprule
\multicolumn{1}{c}{\bf Method}  &\multicolumn{1}{c}{\bf CMNIST}&\multicolumn{1}{c}{\bf CelebA}&\multicolumn{1}{c}{\bf QM9} 
\\ \midrule
\method\ 1 step        & $\mathbf{0.84 \pm 0.02}$ & $\mathbf{0.49 \pm 0.12}$ & $\mathbf{0.29 \pm 0.01}$\\
uncd. 1 step           & $0.022 \pm 0.002$ & $0.02 \pm 0.01$ & $0.19 \pm 0.02$\\
$\tfrac{1}{2}v_t(x|c_1)+\tfrac{1}{2}v_t(x|c_2)$ 1 step            & $0.51 \pm 0.01$ & $0.14 \pm 0.05$ & $0.246 \pm 0.008$\\
only score 1 step      & $0.033 \pm 0.002$ & $0.018 \pm 0.006$ & $0.016 \pm 0.005$\\[2pt]
\method\ 5 step        & $\mathbf{0.912 \pm 0.004}$ & $\mathbf{0.58 \pm 0.09}$ & $\mathbf{0.354 \pm 0.009}$\\
uncd. 5 step           & $0.150 \pm 0.007$ & $0.05 \pm 0.02$ & $0.294 \pm 0.004$\\
$\tfrac{1}{2}v_t(x|c_1)+\tfrac{1}{2}v_t(x|c_2)$ 5 step            & $0.686 \pm 0.004$ & $0.25 \pm 0.05$ & $0.327 \pm 0.006$\\
only score 5 step      & $0.034 \pm 0.001$ & $0.030 \pm 0.002$ & $0.053 \pm 0.007$\\
\bottomrule
\end{tabular}
\end{center}
\end{table}

\subsection{Samples}

\begin{figure}[!h]
    \centering
    \begin{tabular}{@{}cc@{}}
        \includegraphics[width=0.48\textwidth]{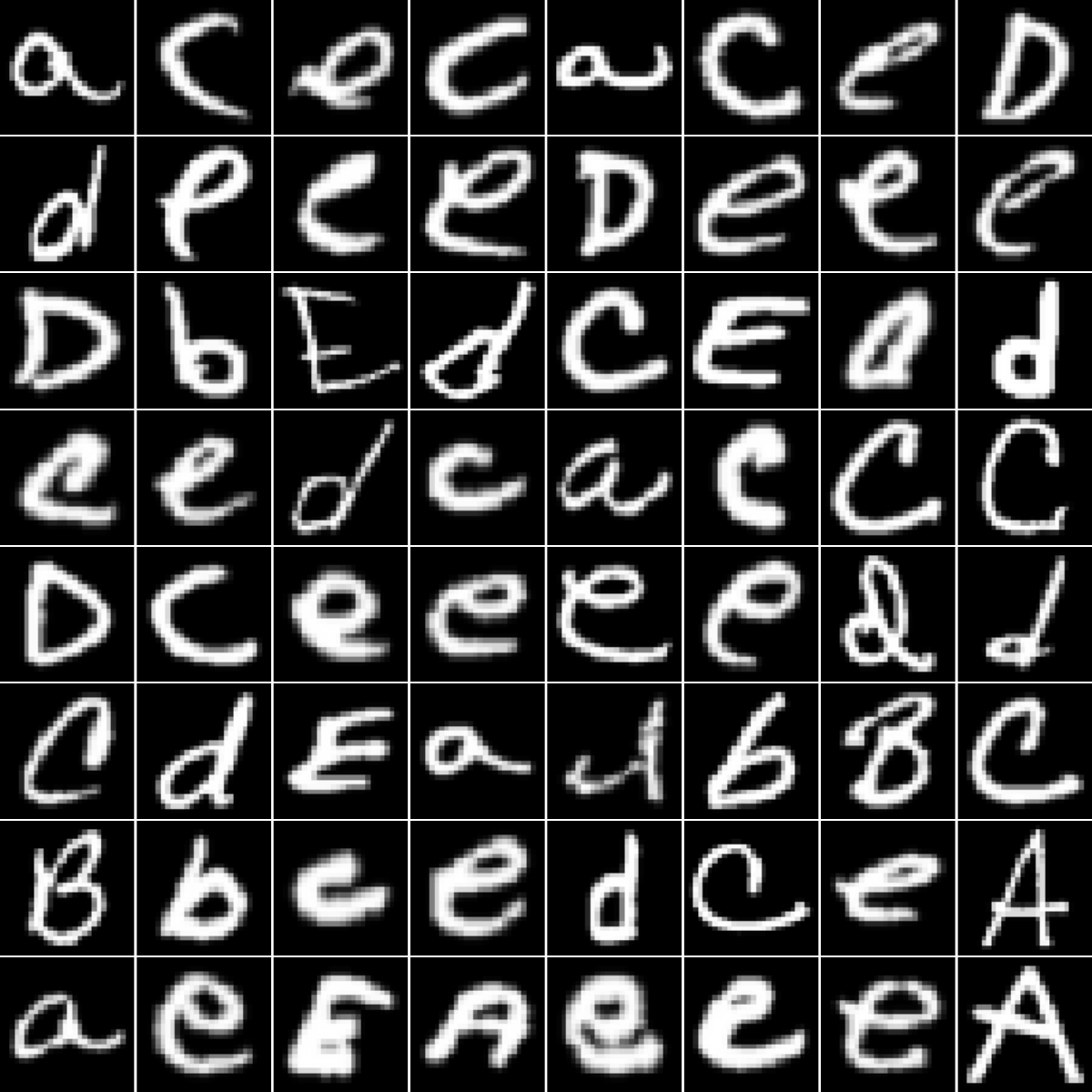} &
        \includegraphics[width=0.48\textwidth]{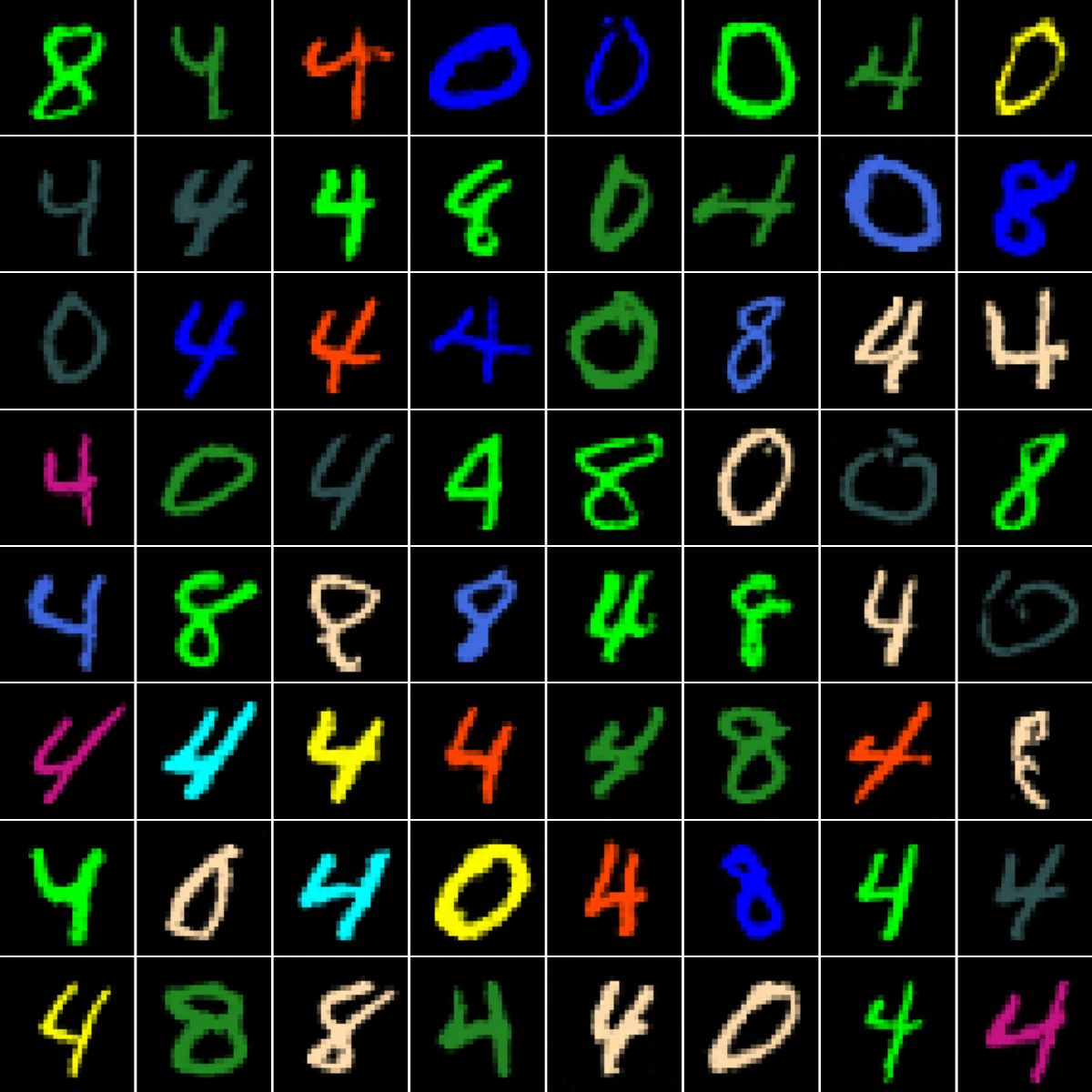} \\
        \includegraphics[width=0.48\textwidth]{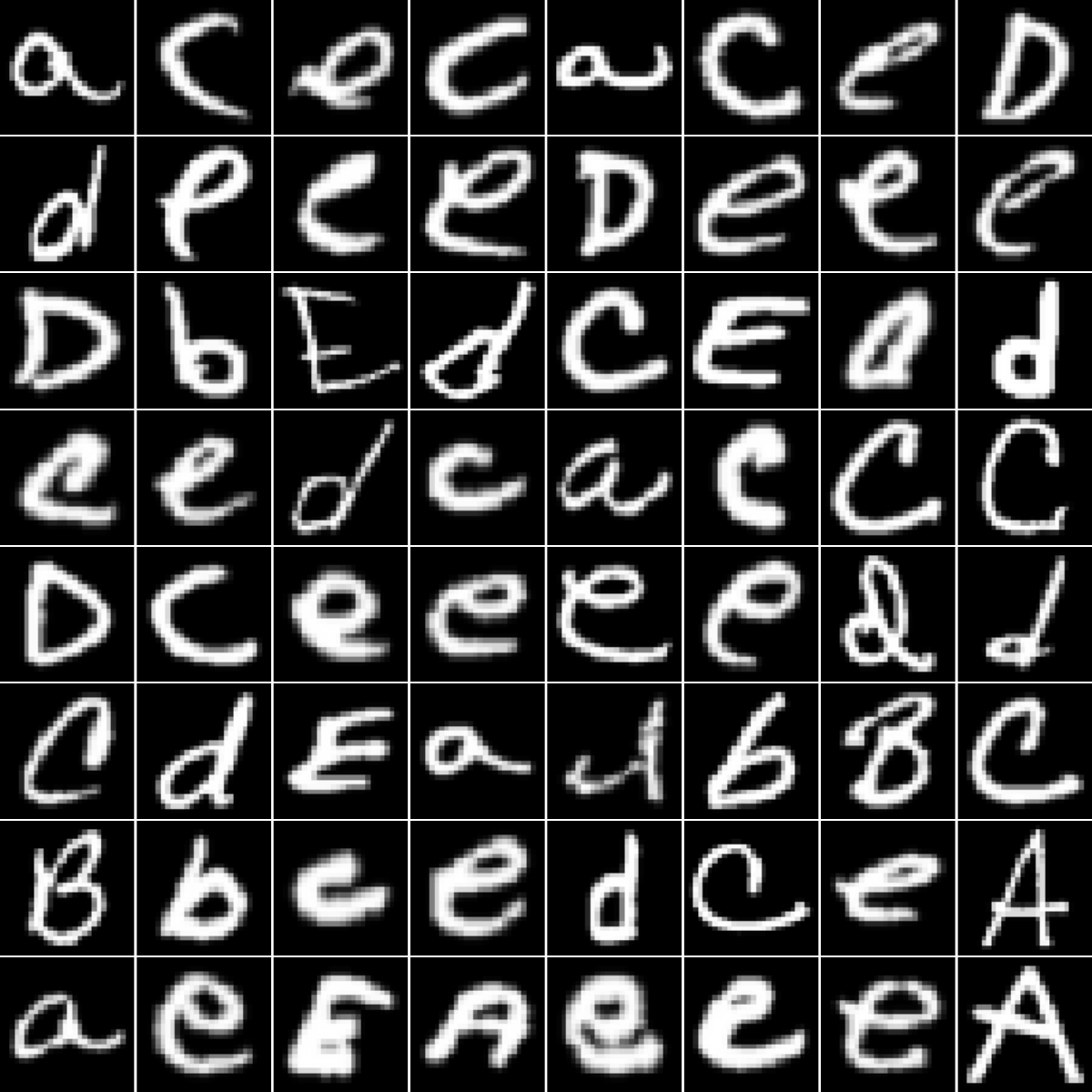} &
        \includegraphics[width=0.48\textwidth]{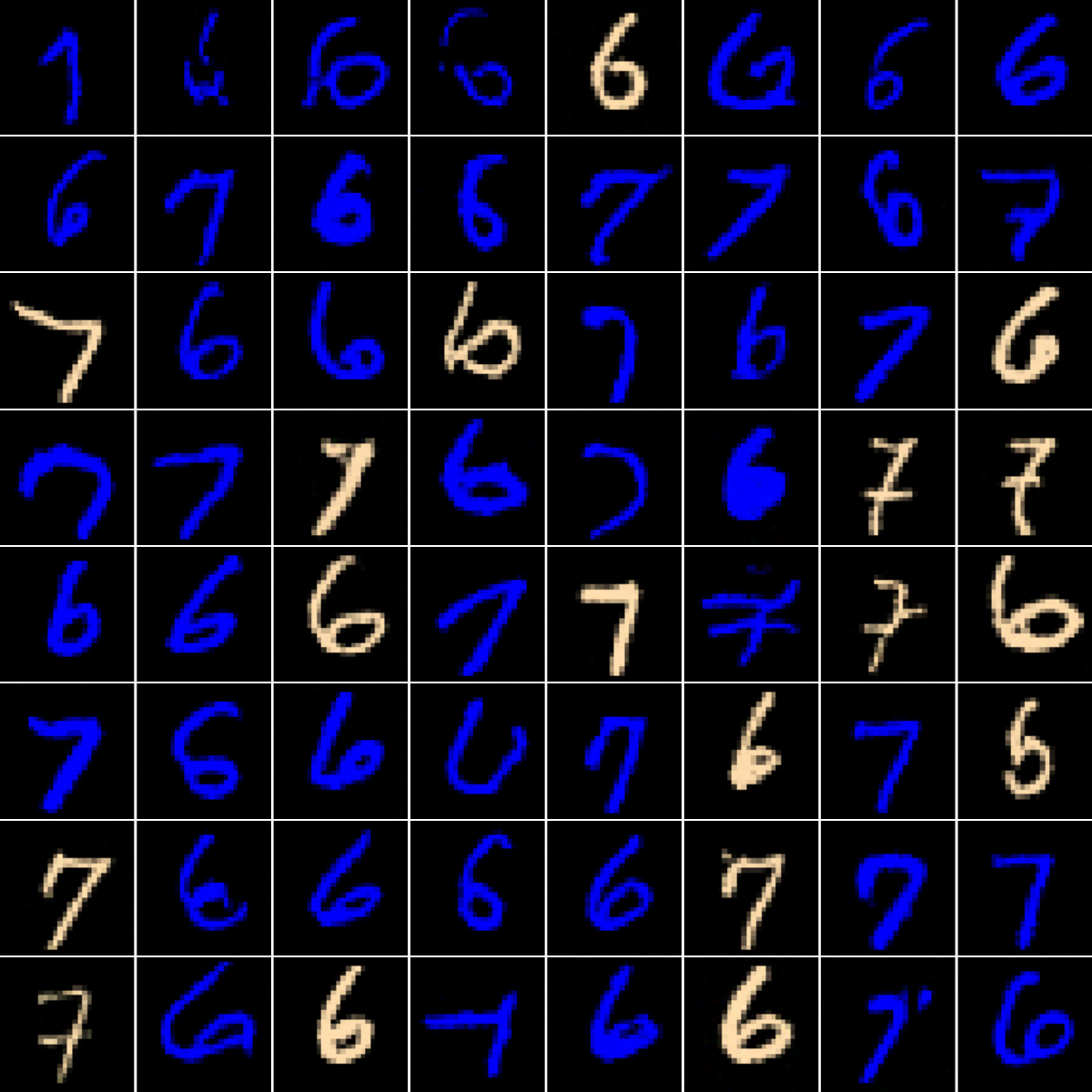}
    \end{tabular}
    \caption{
        \textbf{Samples of the CMNIST experiments.} We plot samples from different randomly generated logical expressions. On the top we use the expression $(8\lor 4)\lor(4\lor 0)$ and on the bottom we use the expression
        $(7\lor6)\land(\texttt{beige}\lor \texttt{blue})$.
    }
    \label{fig:samples_cmnist}
\end{figure}

\begin{figure}[!h]
    \centering
    \includegraphics[width=0.48\textwidth]{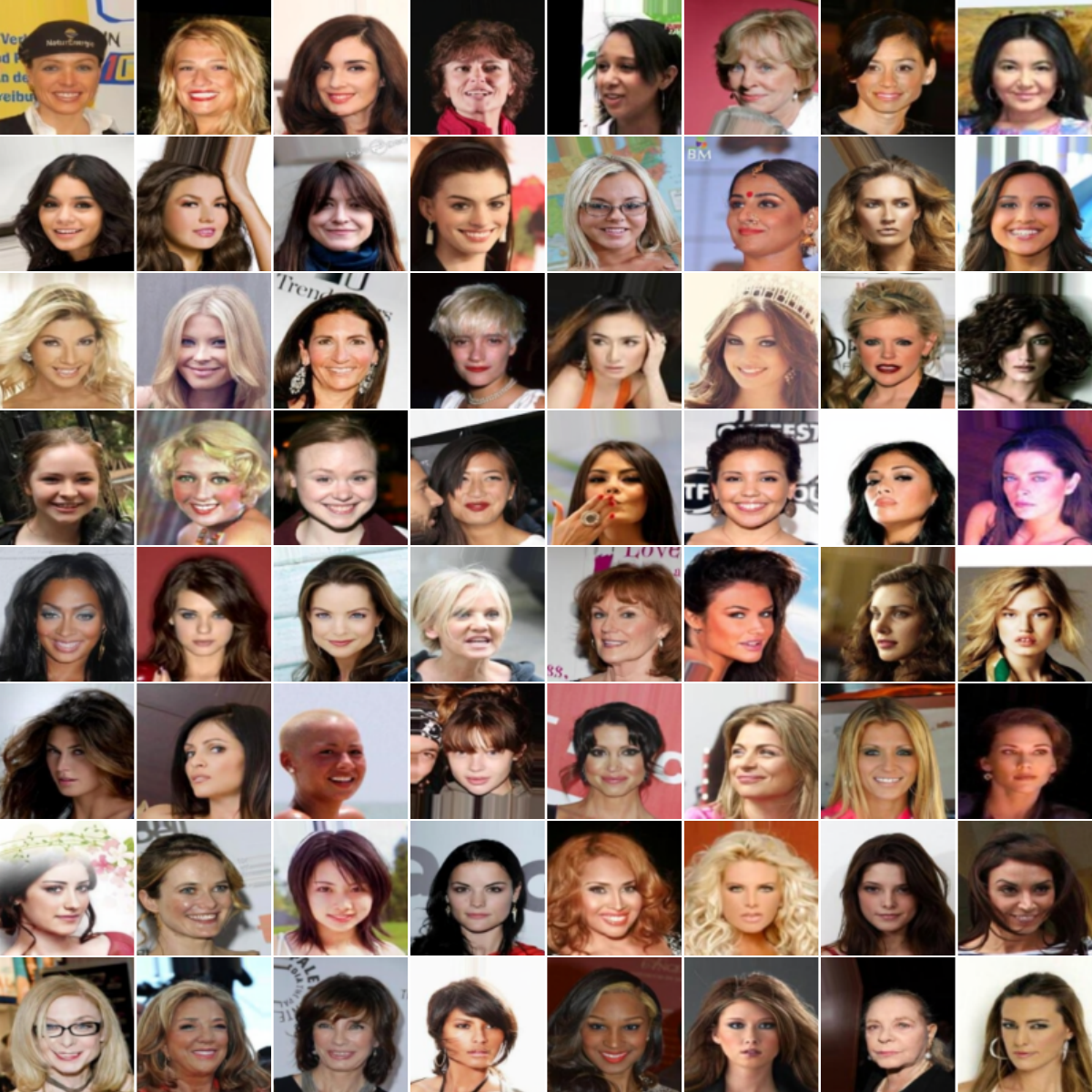}
    \hfill
    \includegraphics[width=0.48\textwidth]{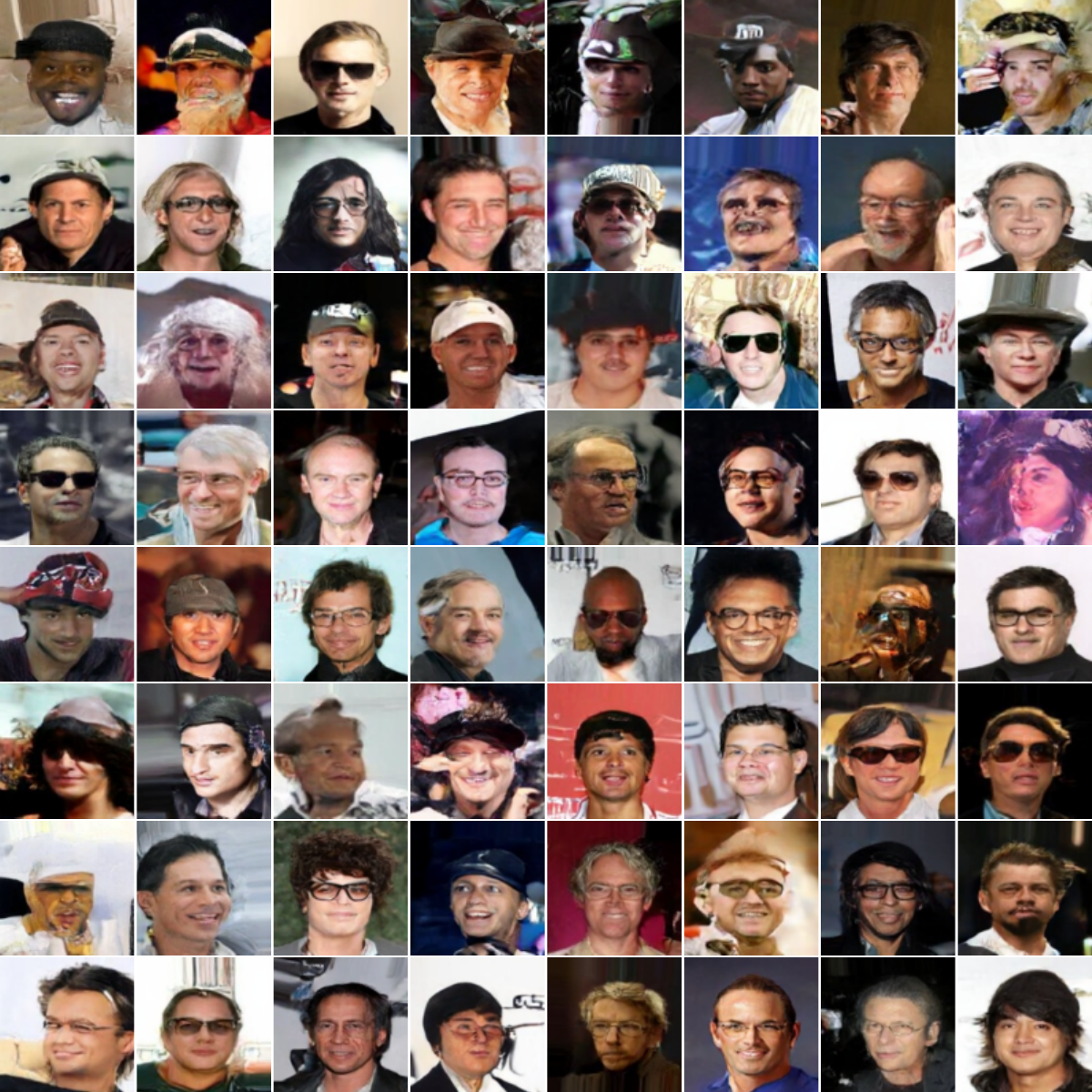}
    \caption{
        \textbf{Samples of the CelebA experiments.} Here we show samples from the logical condition $\texttt{glasses}\lor\texttt{hat}$. On the left the source images and on the right the final images are shown.
    }
    \label{fig:samples_celeba}
\end{figure}

\begin{figure}[!h]
    \centering
    \includegraphics[width=0.48\textwidth]{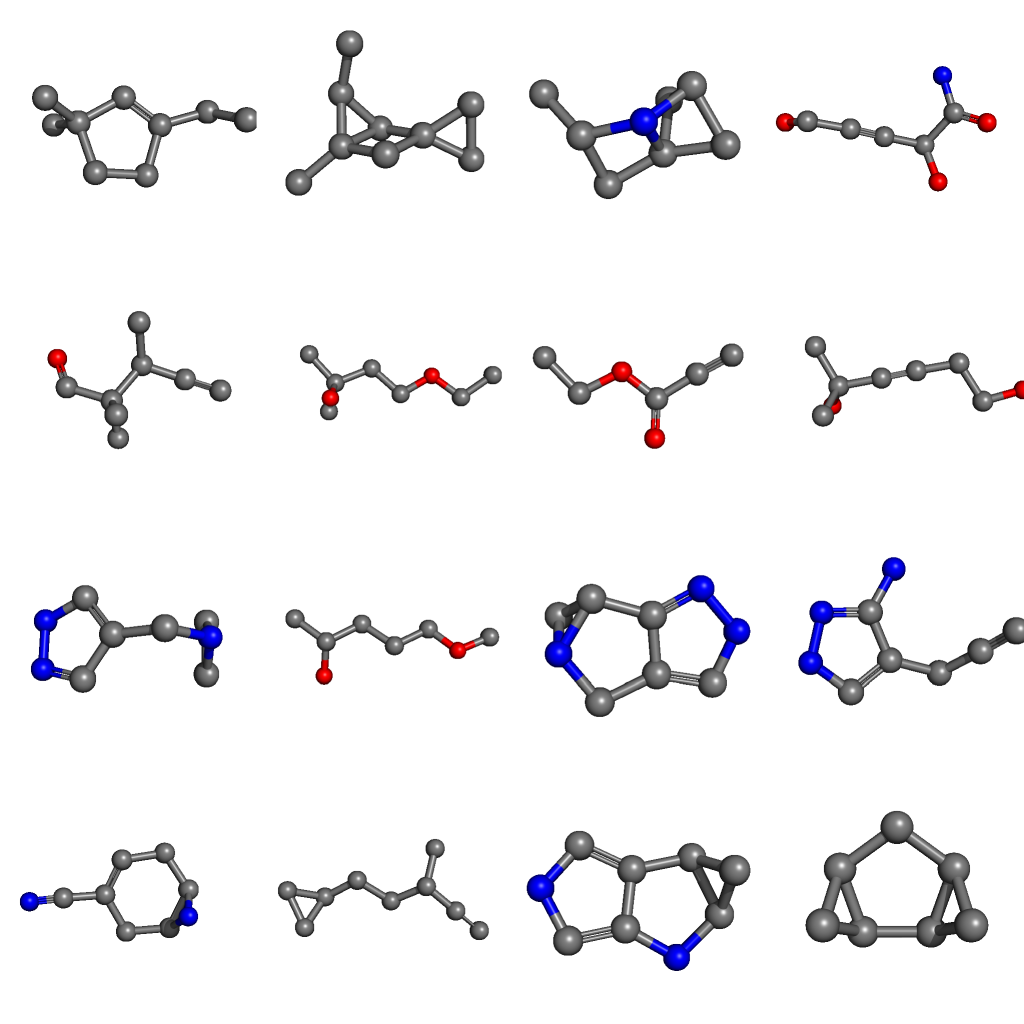}
    \hfill
    \includegraphics[width=0.48\textwidth]{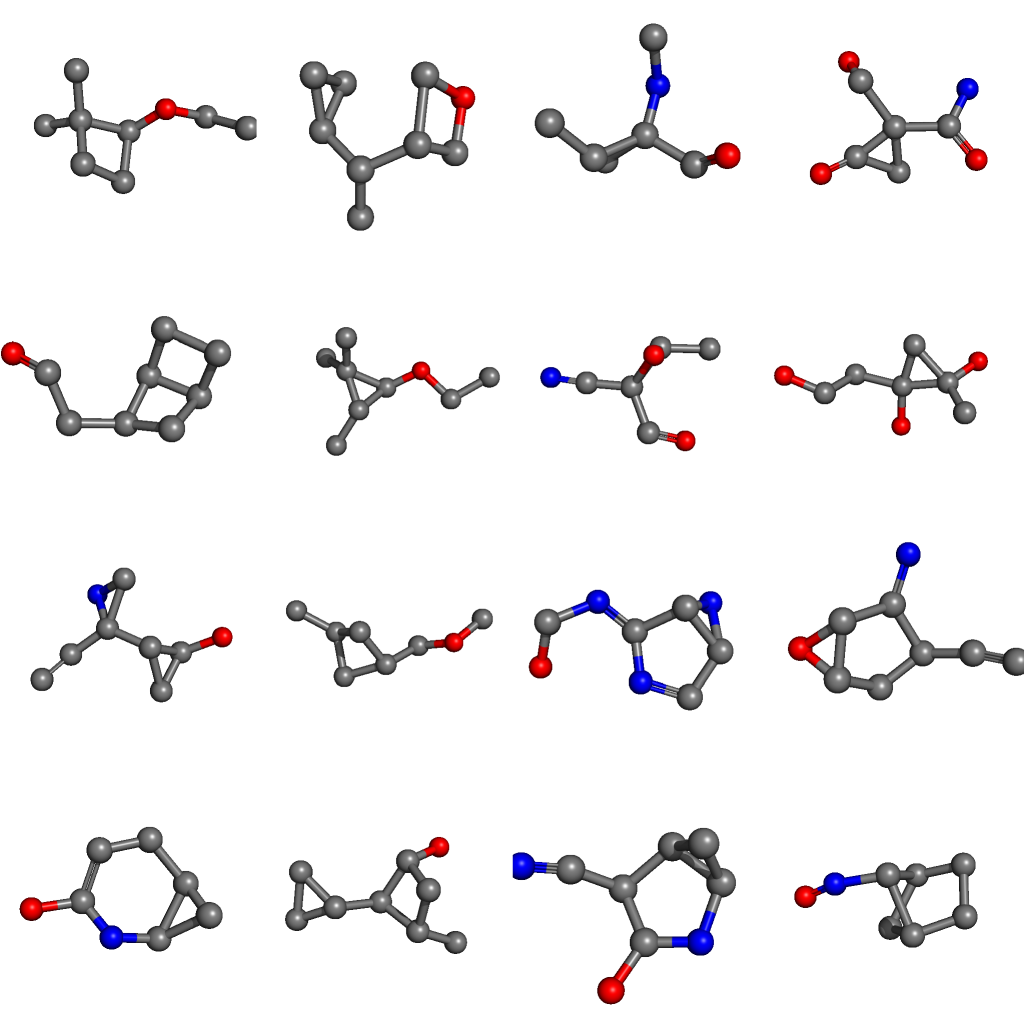}\caption{
        \textbf{Samples of the QM9 experiments.} Both valid and conform samples of the logical statement $\texttt{rings}\land\texttt{oxygen}$ are shown. On the left the starting molecules and on the right the generated molecules are depicted. 
    }
    \label{fig:samples_qm9}
\end{figure}

\end{document}